\documentclass[sigconf]{acmart}

\usepackage{enumitem}
\usepackage{hyperref}       
\usepackage{url}            
\usepackage{booktabs}       
\usepackage{amsfonts}       
\usepackage{nicefrac}       
\usepackage{microtype}      
\usepackage{xcolor}         
\usepackage{graphicx}
\usepackage{subcaption}
\usepackage{ulem}
\usepackage{array}
\usepackage{multirow}
\usepackage{enumitem}
\usepackage{amsmath}
\usepackage{xurl}
\usepackage{xcolor}

\AtBeginDocument{%
  }

\copyrightyear{2026}
\acmYear{2026}
\setcopyright{cc}
\setcctype{by}
\acmConference[KDD '26]{Proceedings of the 32nd ACM SIGKDD Conference on Knowledge Discovery and Data Mining V.2}{August 09--13, 2026}{Jeju Island, Republic of Korea}
\acmBooktitle{Proceedings of the 32nd ACM SIGKDD Conference on Knowledge Discovery and Data Mining V.2 (KDD '26), August 09--13, 2026, Jeju Island, Republic of Korea}
\acmDOI{10.1145/3770855.3817540}
\acmISBN{979-8-4007-2259-2/2026/08}

\begin{document}

\title{GraphChase: A Platform and Benchmark for Urban Network Security Games}


\author{Shuxin Zhuang}
\authornote{These authors contributed equally to this work.}
\orcid{0009-0004-4315-8446}
\affiliation{%
  \institution{City University of Hong Kong}
  \institution{CAIR, Hong Kong Institute of Science \& Innovation, CAS}
  \city{Hong Kong}
  \country{Hong Kong}
}
\email{shuxin.zhuang@my.cityu.edu.hk}

\author{Shuxin Li}
\authornotemark[1]
\orcid{0009-0001-5748-2667}
\affiliation{%
  \institution{Nanyang Technological University}
  \city{Singapore}
  \country{Singapore}
}
\email{shuxin.li@ntu.edu.sg}

\author{Tianji Yang}
\orcid{0009-0007-6678-6244}
\affiliation{%
  \institution{Georgia Institute of Technology}
  \city{Atlanta}
  \state{Georgia}
  \country{United States}
}
\email{tyang425@gatech.edu}

\author{Muheng Li}
\orcid{0009-0001-1588-9792}
\affiliation{%
  \institution{University of Toronto}
  \city{Toronto}
  \state{Ontario}
  \country{Canada}
}
\email{muheng.li@mail.utoronto.ca}

\author{Xianjie Shi}
\orcid{0009-0001-9601-193X}
\affiliation{%
  \institution{The University of Hong Kong}
  \city{Hong Kong}
  \country{Hong Kong}
}
\email{xianjieshi@connect.hku.hk}

\author{Bo An}
\orcid{0000-0002-7064-7438}
\affiliation{%
  \institution{Nanyang Technological University}
  \city{Singapore}
  \country{Singapore}
}
\email{boan@ntu.edu.sg}

\author{Youzhi Zhang}
\authornote{Corresponding author.}
\orcid{0000-0002-2984-734X}
\affiliation{%
  \institution{CAIR, Hong Kong Institute of Science \& Innovation, CAS}
  \city{Hong Kong}
  \country{Hong Kong}
}
\email{youzhi.zhang@cair.cas.org.hk}

\renewcommand{\shortauthors}{Shuxin Zhuang et al.}

\begin{abstract}
    After the achievement of solving two-player zero-sum games, more AI researchers focus on solving multiplayer games. Urban Network Security Games (\textbf{UNSGs}) represent a class of such games, modeling real-world scenarios where law enforcement must strategically allocate limited resources to intercept criminals escaping within urban networks, and have gained considerable research attention. However, progress in this field has been limited by the absence of a standardized experimental platform and realistic benchmarks with heterogeneous travel costs. To address this limitation, we introduce \textbf{GraphChase}, an open-source platform designed to support the development and evaluation of algorithms for UNSGs. GraphChase offers a unified environment for modeling diverse UNSG variants on unweighted and weighted road networks across urban topologies. It also incorporates learning-based algorithms as baseline references for researchers. Furthermore, our experiments with GraphChase reveal that existing approaches to UNSGs still face challenges in terms of robustness and scalability, and suffer performance degradation when deployed under weighted edge costs, highlighting a sim-to-real generalization gap. GraphChase thus provides a realistic testbed for developing and validating UNSGs solvers under realistic travel-time heterogeneity.
\end{abstract}



\maketitle

\section{Introduction}

In urban environments, ensuring public safety and security is crucial for law enforcement agencies. One important issue is the substantial number of innocent bystanders injured or killed during police pursuits \citep{rivara2004motor}. It is therefore essential to develop effective strategies that enable multiple officers to apprehend fleeing criminals while minimizing risks to civilians and property damage. This paper focuses on responding to major incidents such as terrorist attacks or bank robberies, where police officers must swiftly intercept attackers during their escape. Such scenarios demand efficient strategies for criminal apprehension, which can be naturally modeled as UNSGs. In UNSGs, urban road networks are represented as graphs, and multiple pursuers must coordinate their actions to intercept evaders before they escape through the network.

Urban security games have earned considerable attention from researchers due to their practical significance \citep{li2021cfr,xue2021solving,xue2022nsgzero,li2023solving,li2024grasper}. They also instantiate a challenging class of multiplayer games, where agents interact strategically under cooperation, competition, and imperfect information. In the field of AI research, a lot of focus has been placed on computing Nash equilibria \citep{nash1951non,shoham2008multiagent} in two-player zero-sum extensive-form games, where the players receive opposing payoffs \citep{zinkevich2008regret,moravcik2017,brown2018superhuman}. In this scenario, a Nash equilibrium can be computed in polynomial time with respect to the size of the extensive-form game \citep{shoham2008multiagent}, and recent significant achievements, such as achieving superhuman performance in the heads-up no-limit Texas hold'em poker game \citep{moravcik2017,brown2018superhuman}, demonstrate a strong theoretical and practical understanding of this setting. However, computing Nash equilibria in multiplayer games remains an open challenge and is more difficult than in the two-player zero-sum setting \citep{chen20053,zhang2023computing}. Consequently, more and more AI researchers have turned their attention to solving multiplayer games \citep{brown2019superhuman,meta2022human,carminati2022marriage,zhang2023team,mcaleer2023teampsro,zhang2024dag}. Therefore, UNSGs are not only practically important for urban security, but also technically valuable as a testbed for multiplayer decision-making and equilibrium computation.

\begin{figure*}[tbp]
    \centering
\includegraphics[width=2.0\columnwidth]{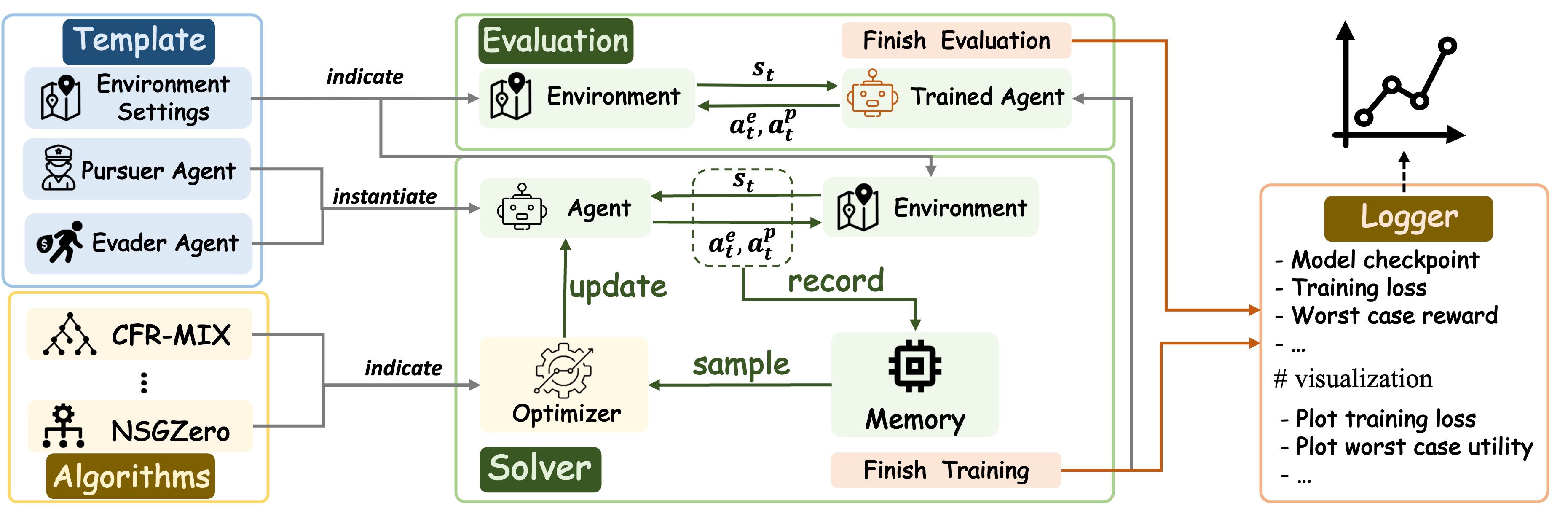}
    \vspace{-8pt}
    \caption{The blueprint of our GraphChase platform.}
    \vspace{-8pt}
    \label{fig:graphchase}
\end{figure*}

However, despite serving as a valuable testbed, solving UNSGs is NP-hard \citep{Jain11,zhang2017optimal,zhang2019optimal}. More specifically, the strategy spaces of players in UNSGs are too large to be enumerated due to the memory constraint \citep{Jain11,zhang2019optimal}. Moreover, when players lack real-time information, they must make decisions under imperfect information. In addition, if police officers are unable to communicate during gameplay, they have to make decisions independently. Finally, UNSGs simultaneously involve cooperation among police officers and competition between the criminal and the team of police officers.
To solve UNSGs, prior works have developed a range of sophisticated algorithms. Specifically, Counterfactual Regret Minimization (CFR) \citep{zinkevich2008regret} has been extended to CFR-MIX \citep{li2021cfr} with deep learning enhancements \citep{brown2019deep}. Additionally, Neural Fictitious Self-Play (NFSP) \citep{heinrich2016deep} has been adapted into NSG-NFSP \citep{xue2021solving} and NSGZero \citep{xue2022nsgzero}. Furthermore, the Policy-Space Response Oracles (PSRO) framework \citep{lanctot2017unified} has also been advanced into Pretrained PSRO \citep{li2023solving} and Grasper \citep{li2024grasper} to improve generalization. All of them are based on the state-of-the-art (SOTA) game-theoretical learning frameworks, powered by deep function approximation.

Although these efforts have achieved some progress in specific problem settings, the UNSGs research community still faces the following challenges. First, \textit{the absence of a unified interface} results in incompatible implementations. Custom environments often return inconsistent data types, data structures, and action spaces. Consequently, an algorithm trained in one environment is often incompatible with others, making cross-evaluation and fair comparison difficult. Second, there is a \textit{realism-scalability trade-off} in environment modeling. While optimization-based methods \citep{zhang2017optimal} incorporate edge weights to model travel times, they rely on mixed-integer linear programming (MILP) and thus struggle to scale to large-scale networks due to computational complexity. Conversely, SOTA learning-based methods typically simplify the environment to unweighted graphs (or uniform costs) to facilitate training. This simplification overlooks the real-world factor: the heterogeneity of road segments (e.g., varying lengths and speed limits) can alter strategies. Neglecting these realistic weights limits the effectiveness of algorithms in real-world scenarios. 

To address these challenges, we introduce an open-source platform, \textbf{GraphChase}, to standardize UNSGs research and enable simulations in realistic environments. GraphChase is designed to provide researchers with a unified framework, as illustrated in Figure~\ref{fig:graphchase}, facilitating both the development and comparison of algorithms.
Specifically, our main contributions are as follows:

\noindent\textbf{1) Development of a Unified UNSGs Platform.} GraphChase provides a standardized interface that decouples the environment from the algorithm. This unification resolves the incompatibility issue, enabling cross-evaluation of different strategies on identical tasks. The platform allows researchers to define custom graph structures and configure key parameters, such as the number and positions of pursuers and evaders. It also includes optimized environment rollout mechanisms for efficient data generation.

\noindent\textbf{2) Establishment of Benchmark Results.} GraphChase integrates several deep learning-based algorithms within a unified game framework, enabling fair comparisons across diverse game configurations. The resulting benchmarks offer reliable baselines for evaluating future algorithms.

\noindent\textbf{3) Revealing the Sim-to-Real Gap.} We conduct extensive experiments on the platform to evaluate benchmark algorithms. The results reveal that current approaches face limitations in scalability and robustness, and suffer performance degradation when deployed on weighted road networks. These findings quantify a sim-to-real generalization gap and establish reference results for future work.

GraphChase lowers the entry barrier to UNSGs research and fosters a community focused on scalable, realistic security challenges.

\section{Urban Network Security Games}

We firstly define urban network security games (UNSGs) to model the interactions between multiple pursuers (police officers) and evaders (criminals) in the urban road networks. Then, building upon prior work on security games in urban environments \citep{Jain11,zhang2017optimal,zhang2019optimal}, we highlight the core challenges involved in solving these games.


\subsection{Game Definition}


Consider the scenario in which pursuers are tasked with capturing evaders escaping through an urban road network, which we formulate as an UNSG on a graph. We represent the road network as a graph $G=(V,E,\omega)$, where $V$ denotes the set of vertices, $E$ denotes the set of edges, and $\omega(u,v)$ denotes the weight of edge $(u,v)\in E$. In $G$, we define a subset of the vertices, $V_{\text{exit}} \subset V$, as exit nodes from which the evader can escape, and let $\mathcal{N}(v)$ denote the set of neighbors of vertex $v$. In UNSGs, graphs may be directed or undirected to represent one-way and two-way streets, and unweighted graphs are treated as a special case by setting $\omega(u,v)=1$ for all edges. This graphical representation provides a flexible way to simulate the game setting. To account for weighted travel costs, agents may be located either at vertices or along edges during movement. Let $l_t^i$ denote the location of agent $i$ at time $t$, represented as a tuple $l_t^i = (u,v,\delta)$, indicating that the agent is situated on the edge $(u,v)$ at a distance $\delta$ from vertex $u$, where $0 \le \delta \le \omega(u,v)$. A vertex position $v$ is represented as $(v,v,0)$.

In UNSGs, the player set includes pursuers and evaders, both modeled as agents moving over the road network $G$. Each side may consist of a single agent or multiple agents.
For example, multiple pursuers may cooperate to capture a single evader or a team of evaders. Formally, $N=(\mathbf{p}, \mathbf{e})$, where $\mathbf{p}=(p_1, p_2,\dots,p_n), n \geq 1$ denotes pursuers and $\mathbf{e}=(e_1, e_2,\dots,e_m), m \geq 1$ denotes evaders. 

Since pursuers can block all exit nodes within a finite time, we model this duration as a lockdown horizon $T$, which also serves as the maximum time horizon of the game. The game proceeds in discrete unit time steps. 
We adopt a hybrid decision-making mechanism, under which an agent makes decisions under two conditions: (1) at the beginning of each time step, all agents observe the information and make decisions simultaneously; and (2) instantaneously upon reaching a vertex during movement execution. 
Therefore, the state-dependent action set $A(l_t^i)$ allows an agent to move to a neighboring vertex when at a vertex, or to continue moving or reverse direction when on an edge. The game proceeds until a termination condition is met. Consequently, the number of decision points can exceed $T$, although the game lasts at most $T$ discrete time steps.
An evader is considered captured when its distance to any pursuer satisfies $d(l_t^{e_j}, l_t^{p_k}) \le \epsilon$, where $d(\cdot)$ represents shortest path distance between the points on the graph and $\epsilon$ is a predefined capture radius. The condition is evaluated at each decision point. Capture is checked independently for each evader; once captured, the evader and its capturing pursuer stop moving, and no release interaction is modeled.
Termination occurs when: (i) \textbf{Capture}: all evaders are captured; (ii) \textbf{Escape}: an evader reaches an exit node $v \in V_{\text{exit}}$; and (iii) \textbf{Timeout}: the game reaches the horizon $T$. 
Upon termination, rewards are assigned based on the outcome: pursuers win in cases (i) and (iii), while evaders win in case (ii).
\subsection{Information and Strategy} 
In different real-world cases, the pursuer and evader may access different information, such as the location information of each player. With the aid of tracking devices, such as the StarChase GPS-based system \citep{gaither2017pursuit}, police officers may get the real-time location of the criminal. In another case, to avoid the worst case, the police officers usually assume that the criminal can get the real-time location of the police officers since they may not know the criminal's ability. Therefore, in our GraphChase platform, we can simulate the following four cases: i) the evader can get the real-time location information of the pursuer while the pursuer cannot get the real-time location information of the evader; ii) the pursuer can get the real-time location information of the evader while the evader cannot get the real-time location information of the pursuer; iii) both the evader and the pursuer can get the real-time location information of the opponent; and iv) both the evader and the pursuer cannot get the real-time location information of the opponent. Another important factor to consider is the communication among pursuers. Specifically, if communication is not allowed during gameplay, each pursuer must make decisions independently. In contrast, if communication is permitted, pursuers can coordinate their actions, and thus may be modeled collectively as a single player in the game.

To define the strategy for each player, we take the case where communication is permitted as an illustrative example. Based on the available real-time location information, the behavior strategy $\sigma_{\mathbf{e}}$ or $\sigma_{\mathbf{p}}$ is defined as a function that maps each decision point to a probability distribution over the corresponding set of available actions. A strategy profile $\sigma$ is then defined as a tuple consisting of one strategy for each player, i.e., $\sigma = (\sigma_{\mathbf{p}}, \sigma_{\mathbf{e}})$.  The pursuer's payoff function is defined by  $u_{\mathbf{p}}(\sigma_{\mathbf{p}}, \sigma_{{\mathbf{e}}})\in \mathbb{R}$, and the game is zero-sum, such that the evader's payoff satisfies $u_{\mathbf{e}}(\sigma_{\mathbf{p}}, \sigma_{{\mathbf{e}}})=-u_{\mathbf{p}}(\sigma_{\mathbf{p}}, \sigma_{{\mathbf{e}}})$. 
We adopt the Nash equilibrium (NE) \citep{nash1950equilibrium} as the solution concept since the NE strategy profile is a steady state in which no player can increase its utility by unilaterally deviating. 
In GraphChase platform, we consider the NE strategy of the pursuer to be the optimal strategy and take the worst-case utility of the pursuer as the measure for the pursuer's strategy, 
i.e., $\mathrm{Worst\text{-}Case\ Utility}_{p} =\max_{\sigma_{\mathbf{p}} \in \Sigma_{\mathbf{p}}} \min_{\sigma_{\mathbf{e}} \in \Sigma_{\mathbf{e}}}u_{\mathbf{p}}(\sigma_{\mathbf{p}}, \sigma_{{\mathbf{e}}}).$


\subsection{Challenges}
\textbf{Computational Limits of Large Strategy Space.} The network-based nature of the environment leads to a strategy space for players in UNSGs that cannot be fully enumerated due to the memory constraints \citep{Jain11,zhang2019optimal}. Specifically, when a player's strategy consists of selecting a path, the number of possible paths in large-scale UNSGs makes exhaustive enumeration infeasible. Even in relatively simple scenarios—such as when time dynamics are ignored, and pursuers can coordinate their actions—solving UNSGs remains extremely challenging \citep{Jain11}. 
Consequently, we could expect even greater difficulties in more complex settings.

\noindent\textbf{Partial Observation and Imperfect Information.} UNSGs typically feature imperfect information and asymmetric knowledge among participants. For example, pursuers may not have full knowledge of the evader’s location or intended actions, while the evader may also possess only limited information about the pursuers. Furthermore, partial observability and uncertain time horizons further complicate the decision-making process. These difficulties highlight the need for robust algorithms capable of operating effectively under conditions of uncertainty.

\noindent\textbf{Competition and Cooperation among Agents.} UNSGs involve adversarial competition between pursuers and the evader, where the success of one side corresponds to the loss of the other. At the same time, pursuers cooperate within their team under a shared utility function to maximize the probability of capturing the evader. This intrinsic interplay between competition and cooperation increases overall complexity and calls for algorithmic solutions that balance optimal adversarial strategies with effective team coordination.

These challenges make UNSGs an ideal testbed for evaluating algorithms in complex and dynamic environments. 
The lack of a unified platform in UNSGs research motivated us to develop GraphChase as a standardized benchmark for future studies.


\begin{figure}[tp]
  \centering
  \includegraphics[width=\columnwidth]{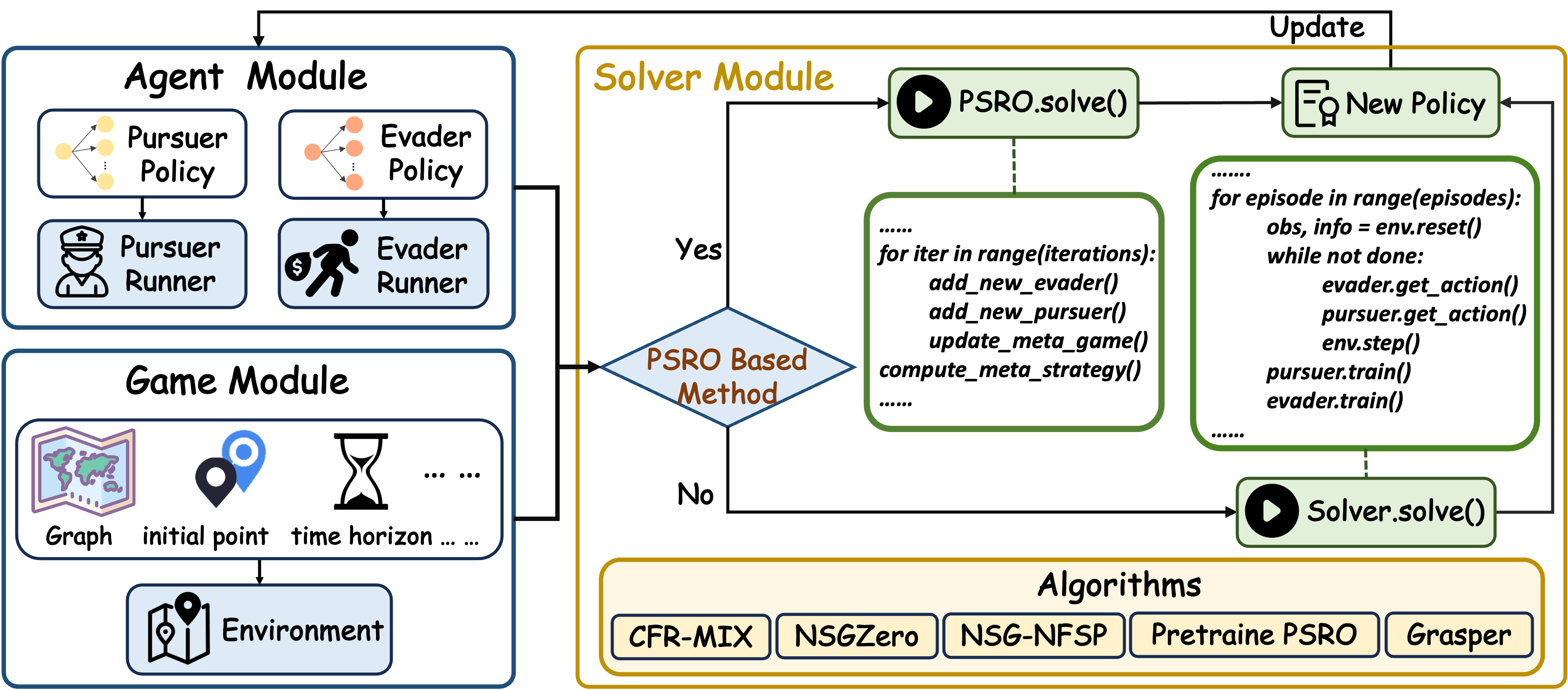}
  \caption{The core structure and workflow of GraphChase.}
  \label{fig:game}
\end{figure}

\section{Platform: GraphChase}
Figure \ref{fig:graphchase} outlines the setup workflow of GraphChase. After user configuration, the platform executes training and evaluation procedures. We next describe the core components of GraphChase, as shown in Figure~\ref{fig:game}. A usage guide is available in Appendix~\ref{usage}.

\subsection{Core Components}
GraphChase is designed to support the construction of diverse graph structures and allow custom algorithms to be implemented for solving various UNSGs. To achieve this, the platform adopts a modular architecture that decouples its core functions into three main components: Game (environment), Agent (simulated agents), and Solver (algorithm module). Users can define custom problem settings, use built-in baseline algorithms by the platform, or implement their own methods, thereby improving research efficiency. 
 
\textbf{Game Module.} The Game module implements the UNSG environment and supports the following features:

(1) \textit{Graph Structure Definition}: GraphChase represents road networks as NetworkX graphs. Users can construct graphs from an adjacency matrix or import a NetworkX \texttt{gpickle} file.

(2) \textit{Basic Parameter Configuration}: Game parameters, such as the number of pursuers and evaders, initial positions, exit nodes, and the maximum time horizon, can be specified during initialization. Details of the configurable parameters are provided in Appendix~\ref{controllparams}.

(3) \textit{Weighted Graphs and State Representation}: GraphChase supports weighted road networks by associating each edge $(u,v)\in E$ with a weight $\omega(u,v)>0$, representing the continuous travel time required to traverse that edge. Edge weights are either loaded from the NetworkX edge attribute or constructed from a provided adjacency matrix. To account for varying edge weights within the discrete-time game, we model agent locations in continuous space. Each agent $i$ is represented by a tuple $x^i=(u, v, \delta)$, where $u, v \in V$ denote the endpoints of the current edge and $0 \le \delta \le \omega(u, v)$ is the distance from vertex $u$. A node position at vertex $v$ is encoded as $(v, v, 0)$. The maximum time horizon $T$ serves as a global temporal constraint in the same units as the edge weights. Each movement consumes a continuous time budget equivalent to the weight on the edges. A Timeout occurs if the cumulative travel time reaches $T$ before capture or escape conditions are met. 

(4) \textit{Continuous-Time Step with a Unit Budget}: GraphChase uses a continuous-time transition model with a fixed time budget of $1.0$ per environment step. In each call to \texttt{Env.step()}, the simulator advances by a duration $\Delta t \in(0,1]$, where $\Delta t$ is limited by the remaining time budget in the current step. All agents are updated synchronously: they move for $\Delta t$ along their chosen directions, and for an agent $i$ on an edge, its remaining distance $d^i$ decreases by $\Delta t$. If $d^i$ becomes $0$, the agent reaches the endpoint and is encoded as $(v,v,0)$. As a result, traversing an edge with weight $\omega(u,v)>1$ requires multiple environment steps.

(5) \textit{Actions, Termination, and Observations}: Actions are node IDs (start from 1) with a reserved action $0$ for \emph{stay}. If an agent is at a node $v$ (i.e., $(v,v,0)$), it may choose any neighbor $w\in\mathcal{N}(v)$ (or stay) to start traversing $(v,w)$. If an agent is on an edge $(u,v)$, it may choose to continue toward the destination, reverse direction toward the start node, or stay, corresponding to actions in $\{0,u,v\}$. GraphChase employs a distance-based criterion for termination. An evader is considered caught if the shortest path distance to any pursuer satisfies $d(\cdot) \le \epsilon$, where $\epsilon$ is a predefined capture radius. Escape is achieved if all evaders reach target exit nodes $v \in V_{\text{exit}}$. To accommodate continuous edge movement, we adopt a hybrid checking strategy, evaluating capture and escape conditions both at discrete time steps and whenever an agent reaches a vertex. The episode terminates upon Capture (all evaders are caught), Escape, or Timeout (the maximum time horizon $T$ is reached). The environment provides a joint observation containing the full states of both sides (including specific edge positions); users can implement alternative information structures by wrapping the environment or customizing the observation function.

\textbf{Agent Module.} The Agent module provides reusable implementations of agent policies and environment-interaction runners. It is designed to decouple (i) \emph{policy representation} (neural or non-neural), (ii) \emph{trajectory collection
and training pipelines}, and (iii) \emph{optimization logic} (delegated to the Solver module). It revolves around two core abstractions: \texttt{Policy} and \texttt{Runner}.

(1) \textit{Unified Policy Interface:} The module defines a lightweight interface for action selection and value estimation. This design allows solvers to treat heterogeneous policies uniformly---whether they are deep neural networks (e.g., PPO) or structured heuristics (e.g., path-search algorithms). This flexibility is critical for UNSGs, where baselines mix learning-based and rule-based strategies. Beyond the path-based policies used by UNSG solvers, this interface also supports step-wise decision policies that act on environment observations at each decision point, enabling neural and heuristic policies to be implemented under the same environment interface.

(2) \textit{Runner-Based Abstraction:} To decouple the solver from low-level simulation loops, we introduce \texttt{Runners} to encapsulate environment interaction, preprocessing, and batch construction. Runners handle observation encoding and action masking, exposing only processed tensors to the agents. This design also supports vectorized rollouts, significantly accelerating data collection.

(3) \textit{Support for Iterative Game-Theoretic Learning:} The module targets solvers like PSRO. It supports policy cloning and persistence, enabling policy population creation and best response training without re-implementing environment logic or architectures.

\textbf{Solver Module.} The Solver module implements a suite of learning based and game-theoretic solvers for UNSGs, and is responsible for (i) updating agent parameters through reusable optimization algorithms and (ii) orchestrating higher-level equilibrium-finding procedures (e.g., PSRO) that iterate between best response computation and meta-strategy updates. It is decoupled from the game mechanics, interacting with agents solely via the standardized Runner interfaces. The Solver module supports the following features:

(1) \textit{Plug-and-Play Optimizers}: GraphChase provides reusable optimizers such as PPO and MAPPO for best response training. These algorithms consume batches prepared by runners and update agent parameters without interacting with the environment. Consequently, a new environment or a new policy network can reuse the optimizer as long as the runner outputs the expected batch fields (e.g., action indices, log-probabilities, rewards, and value targets).

(2) \textit{Support for Game-Theoretic Learning (PSRO)}: GraphChase supports Policy Space Response Oracles (PSRO). The framework decomposes PSRO into modular components: (i) a \textbf{Best Response Oracle} that can leverage either RL (e.g., PPO and MAPPO) or heuristic search, and (ii) a \textbf{Meta-Solver} (e.g., Projected Replicator Dynamics) that operates on payoff matrices. This modularity enables researchers to apply the PSRO framework via the GraphChase interface, eliminating redundant code. It also allows researchers to experiment with custom meta-solvers and best response oracles.

(3) \textit{Support for Custom Solvers}: Beyond baselines like PSRO and other benchmark algorithms, GraphChase supports custom algorithm implementation. Users can define their own solvers via our unified interface without modifying the underlying game core.

\subsection{Benchmark Protocol and Evaluation}
\label{sec:BenchmarkAlgorithms}

GraphChase provides a standardized benchmark protocol with fixed evaluation instances, a canonical evader execution protocol, and unified evaluation metrics. We next describe the benchmark algorithms and evaluation methods.

To establish a benchmark for researchers, GraphChase integrates several representative algorithms for UNSGs. (1) \textbf{CFR-MIX} algorithm \citep{li2021cfr} enhances traditional counterfactual regret minimization (CFR) \citep{zinkevich2008regret} with deep learning techniques \citep{brown2019deep}. (2) \textbf{NSG-NFSP} \citep{xue2021solving} adapts the neural fictitious self-play method \citep{heinrich2016deep} by producing action-value outputs for the best response policy and transforming the average policy into a classifier that assigns distributions only over valid actions. (3) \textbf{NSGZero} \citep{xue2022nsgzero} develops deep neural networks to conduct neural Monte Carlo tree search, supporting efficient joint network training. Variants of the PSRO framework include: (4) \textbf{Pretrained PSRO} \citep{li2023solving} initially trains the pursuer’s model against a variety of evader strategies before refining it with the PSRO loop to improve best response performance; and (5) \textbf{Grasper} \citep{li2024grasper} leverages a graph neural network to represent the graph structure as hidden vectors and utilizes a hypernetwork to generate pursuer policies from these representations. It then follows a three-phase process: pre-pretraining for robust game encoding, multi-task pre-training for stabilizing policy learning, and PSRO-based fine-tuning.

\textbf{Evaluation.} In UNSGs research, evaders make decisions based on valid paths or source-target pairs. Specifically, before the pursuit-evasion process begins, the evader selects one target from a set of feasible exits. The evader then chooses a path from its initial position to the selected target and follows this predetermined path at each time step. The benchmark algorithms adopt this decision-making approach for the evader. As such, our canonical benchmark experiments for existing UNSG solvers are conducted under this path-based execution protocol. In addition, GraphChase also supports step-wise policy execution, where agents sample actions from their policies at each decision point. This execution mode can be instantiated by neural policies such as PPO or by rule-based heuristic policies, without changing the evaluation metrics. Therefore, the primary objective remains to evaluate the optimal strategy for the pursuer. GraphChase provides three evaluation methods.
\begin{itemize}[left=0pt,labelsep=5pt]
    \item \textbf{Worst-Case Utility.} This method enumerates all paths the evader might take and evaluates pursuer utility along each one. The pursuer's utility on the path that yields the lowest value is taken as the worst case. This provides a measure of the pursuer's performance under the most disadvantageous evasion strategy.
    \item \textbf{Pseudo Worst-Case Utility.} As the size of the graph increases, exhaustive enumeration of all evader paths becomes time consuming. To address this, we use pseudo worst-case utility as an approximation to the worst-case exit setting. For each target exit, several paths leading to this exit are randomly sampled, and the average pursuer utility over these paths is used to estimate the pursuer's expected utility conditioned on that exit. We take the minimum value across all exits as the pseudo worst-case utility. 
    \item \textbf{Visualization.} GraphChase supports process visualization. Visualizing trajectories and decision-making processes allows researchers to assess whether a strategy aligns with human intuition. An example is provided in Appendix~\ref{sec:visualization}. 
\end{itemize}


\begin{figure*}[tbp]
    \centering
    \includegraphics[width=0.95\textwidth]{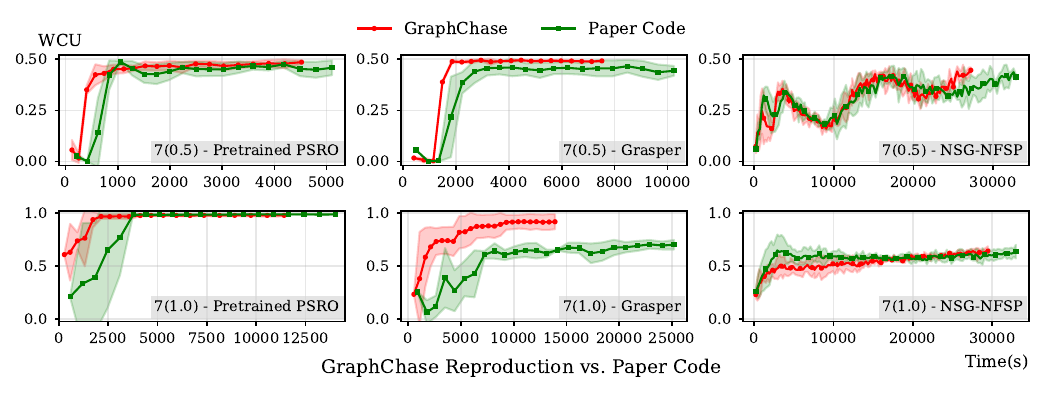}
    \caption{Pursuer pseudo worst-case utility curves during training on the $7\times 7$ grid graph. Each figure title specifies the game configuration and the algorithm. The x-axis denotes the training duration in seconds. Solid lines represent the mean worst-case utility, while shaded regions indicate the standard deviation calculated across 3 independent seeds. 
    }
    \label{fig:exp7_7_reproduce}
\end{figure*}

\section{Experimental Evaluation}

In this section, we evaluates the correctness, efficiency, and versatility of GraphChase. We use the platform to benchmark SOTA algorithms on various real-world maps and analyze their limitations regarding robustness and scalability. Throughout the experiments in this section, we report pseudo worst-case utility and denote it as WCU for brevity. All experiments were performed on a server with a 48-core Intel Xeon Gold 6248R CPU and 8 NVIDIA A30 GPUs\footnote{Our code is available at https://github.com/GraphChase/GraphChasePlatform.git}.

\subsection{Reproducibility and Correctness Validation}
\label{sec:exp1}
A primary objective of GraphChase is to ensure that implemented algorithms reproduce results from their original literature\footnote{Codes were shared by the original authors of these algorithms.}. To verify this, we evaluate Pretrained PSRO, Grasper, NSGZero, NSG-NFSP, and CFR-MIX on undirected grid graphs of $5\times 5$ and $7\times 7$. Each setting involves 4 pursuers and 1 evader. By designing exit locations and agent starting positions, we established two difficulty levels per graph size, corresponding to ground-truth capture probabilities (equivalently, the pursuer’s worst-case utility) of 0.5 and 1.0. See Appendix~\ref{appendix_games} for how these ground-truth values are computed. These settings are denoted by their grid size and probability, where $7(0.5)$ represents a $7\times 7$ grid with a capture probability of 0.5.

Figure \ref{fig:exp7_7_reproduce} illustrates the pursuer's worst-case utility for Pretrained PSRO, Grasper, and NSG-NFSP against wall-clock training time on the $7\times 7$ grid. Using the same hyperparameters as the original studies, our implementations in GraphChase exhibit convergence trajectories that align with the original baselines. GraphChase resolves inefficiencies in the original Grasper codebase, particularly the partial use of collected batch data during updates, resulting in better training performance while maintaining consistent logic. 

Table \ref{tab:large_reproduce} summarizes the converged performance for all five algorithms across the four game settings. The final worst-case utility values are generally consistent with those reported in the original codebases, except for Grasper. This consistency across different graph scales and difficulty levels validates that GraphChase replicates the performance of the reference implementations. The relatively weak performance of CFR-MIX on the $7 \times 7$ settings may be attributed to its product-form decomposition of the team strategy space, which restricts joint pursuer strategies to decomposable individual strategies and can limit expressiveness when stronger team coordination is required. The detailed structures and visualizations of the maps used in our experiments can be found in Appendix~\ref{app:graph_structure_visualization}.

\begin{table*}[t]
\centering
\setlength{\tabcolsep}{5pt} 
\begin{tabular}{l|lcccc}
\toprule
\multicolumn{1}{l}{Algorithm} & Method & $7(0.5)$ & $7(1.0)$ & $5(0.5)$ & $5(1.0)$ \\
\midrule

\multirow{2}{*}{Pretrained PSRO} & GraphChase
& $0.4855 \pm 0.0055$ & $0.9803 \pm 0.0008$ & $0.4928 \pm 0.0088$ & $0.9752 \pm 0.0012$ \\
& Paper Code
& $0.4535 \pm 0.0040$ & $0.9891 \pm 0.0007$ & $0.4740 \pm 0.0026$ & $0.9891 \pm 0.0001$ \\
\midrule

\multirow{2}{*}{Grasper} & GraphChase
& $0.4845 \pm 0.0028$ & $0.9779 \pm 0.0017$ & $0.4547 \pm 0.0059$ & $0.9515 \pm 0.0025$ \\
& Paper Code
& $0.4446 \pm 0.0068$ & $0.7857 \pm 0.0028$ & $0.4147 \pm 0.0115$ & $0.9691 \pm 0.0033$ \\
\midrule

\multirow{2}{*}{NSGZero} & GraphChase
& $0.3268 \pm 0.0571$ & $0.9997 \pm 0.0003$ & $0.4673 \pm 0.0233$ & $0.9999 \pm 0.0001$ \\
& Paper Code
& $0.3375 \pm 0.0293$ & $0.9927 \pm 0.0004$ & $0.4634 \pm 0.0295$ & $0.9876 \pm 0.0003$ \\
\midrule

\multirow{2}{*}{NSG--NFSP} & GraphChase
& $0.4357 \pm 0.0086$ & $0.6374 \pm 0.0060$ & $0.4415 \pm 0.0364$ & $0.8457 \pm 0.0154$ \\
& Paper Code
& $0.4243 \pm 0.0097$ & $0.6286 \pm 0.0077$ & $0.4447 \pm 0.0237$ & $0.7803 \pm 0.0237$ \\
\midrule

\multirow{2}{*}{CFR--MIX} & GraphChase
& $0.212 \pm 0.0281$ & $0.225 \pm 0.0106$ & $0.4167 \pm 0.0219$ & $0.5497 \pm 0.0168$ \\
& Paper Code
& $0.217 \pm 0.0384$ & $0.229 \pm 0.0201$
 & $0.4115 \pm 0.0287$
 & $0.5354 \pm 0.0206$ \\
\bottomrule
\end{tabular}
\caption{Comparison of converged pursuer pseudo worst-case utility across four game settings. Experiments were conducted using identical hyperparameters to the original baselines. For each run, the utility is averaged over the stable convergence phase; the reported values present the aggregate mean $\pm$ standard deviation across three different random seeds.}
\label{tab:large_reproduce}
\vspace{-10pt}
\end{table*}

\subsection{Computational Efficiency Analysis} 
Figure \ref{fig:exp7_7_reproduce} shows that under the same hyperparameters, GraphChase reduces time overhead for training process. Using Pretrained PSRO as a case study, we compared the efficiency of GraphChase against original implementations on the $7(0.5)$ setting, with identical hyperparameters and the same GPU/CPU setup, across three metrics: the time to compute the Evader Best Response (BR), Pursuer BR, and sampling throughput measured in Episodes Per Second (EPS). 

\begin{figure}[!t]
    \centering
    \includegraphics[width=0.95\columnwidth]{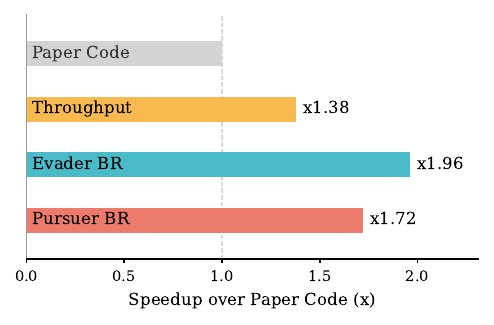}
    \caption{GraphChase speedup over the paper code across throughput and best response computation. The throughput ratio divides the EPS of GraphChase by the EPS of the paper code. Similarly, the speedup for best response computation is the ratio of paper-code execution time to GraphChase execution time. The dashed vertical line indicates the paper code baseline ($\times 1.0$); values larger than 1 denote faster execution.}
    \label{fig:speedup_comparison}
\end{figure}

As shown in Figure~\ref{fig:speedup_comparison}, GraphChase improves the overall sampling throughput by $1.38\times$, indicating more episodes can be generated per second under the same hardware. Moreover, GraphChase accelerates best response computation on both sides: the evader BR is sped up by $1.96\times$ and the pursuer BR by $1.72\times$. This efficiency stems from the vectorized environment design and optimized graph operations, making the platform suitable for large-scale training. More details can be found in Appendix~\ref{convergefaster}.

\subsection{Real-World Urban Topology Benchmarks} 

To evaluate the generalization capabilities of current UNSG algorithms, we established a leaderboard using real-world urban topologies provided by GraphChase. We selected six distinct maps of varying scales and complexities: Singapore, Manhattan, Mumbai, Scotland Yard, Times Square, and the Sydney Opera House. These topologies were sourced from \textit{OpenStreetMap} and abstracted into graph structures. We modeled these topologies as unweighted, undirected graphs to align with the existing state-of-the-art algorithms and ensure a fair evaluation baseline.

Table~\ref{tab:real_world_benchmark} presents the benchmark results across real-world maps. The results reveal performance variance depending on the topology. While most algorithms achieve high pursuer worst-case utility, the utility values are lower in environments such as Manhattan and Times Square, which are characterized by higher numbers of nodes, edges, and extended time horizons. These results provide a baseline for evaluating UNSG solutions on realistic networks.


\subsection{Robustness Evaluation}

In addition to standard performance metrics, we evaluate the robustness of learned policies against environmental shifts.

\textbf{Robustness to Horizon Shift.}
We investigate the robustness of policies to changes in the max time horizon ($T$). Pursuer policies were trained on $7(0.5)$ and $7(1.0)$ settings with $T=4$, then frozen and tested at an extended horizon of $2T$. As shown in Table~\ref{tab:robustness}, the pursuer’s worst-case utility decreases under horizon extension, indicating limited robustness to horizon shifts.

\textbf{Robustness to Edge Weight Variations.}
Motivated by urban dynamics, we examine robustness against structural variations. While the topology of a network remains static, traffic conditions change, altering the travel time of edges. 
To simulate cost variations, we apply static reweighting by randomly assigning travel-time costs to edges and using symmetric weights ($w(u,v)=w(v,u)$), fixed within each episode.
We test policies trained on unweighted graphs under the weighted setting. As shown in Table~\ref{tab:undirected_weighted}, the pursuer’s utility drops significantly in the weighted setting, thereby underscoring limited robustness to edge-cost heterogeneity.

\subsection{Scalability Evaluation}

To evaluate the scalability of existing algorithms, we apply benchmark methods to solve larger game instances. Specifically, we conduct experiments on a $100\times 100$ grid graph with a maximum time horizon of $T=200$. In this setting, four pursuers aim to capture a single evader, who seeks to evade capture by escaping through one of twelve exit nodes. We observed that none of the algorithms were able to complete the training process despite running the experiments for several days. In practice, even on a $30\times 30$ grid, after two days of training, the run still stalled at the path-enumeration stage.

This failure stems from exhaustive enumeration of the evader action space. These methods require pre-calculating all simple paths from the evader's starting position to the exits within the maximum time horizon $(T)$ to facilitate decision-making during simulations. As the graph size and time horizon increase, the number of valid paths grows exponentially leading to computational intractability. In our scalability tests, the computational overhead becomes substantial once the grid size exceeds $20\times20$, and the path-enumeration stage can dominate the overall running time before training proceeds. This exponential explosion in path enumeration time constitutes the scalability barrier for current UNSG solvers and suggests that new approaches must move away from global action space traversal. More details can be found in Appendix~\ref{app:scalability_analysis}.

\begin{table*}[!t]
\centering
\begin{tabular}{lccccccc}
\toprule
& Grasper & Pretrained PSRO & NSGZero & NSG-NFSP & Nodes & Edges & T \\
\midrule
Manhattan      & $0.5897 \pm 0.0540$ & $0.1311 \pm 0.0033$ & $0.0057 \pm 0.0003$ & $0.4108 \pm 0.0933$ & 620 & 1081 & 13 \\
Mumbai         & $0.9923 \pm 0.0009$ & $0.9267 \pm 0.0188$ & $0.9981 \pm 0.0011$ & $0.7008 \pm 0.0302$ & 71  & 92   & 8  \\
Singapore      & $0.9972 \pm 0.0003$ & $0.9254 \pm 0.0100$ & $0.9982 \pm 0.0001$ & $0.8004 \pm 0.0192$ & 158 & 212  & 12 \\
Scotland Yard  & $0.9876 \pm 0.0013$ & $0.1573 \pm 0.0032$ & $0.9981 \pm 0.0007$ & $0.5531 \pm 0.0390$ & 200 & 391  & 9  \\
Sydney Opera House   & $0.9956 \pm 0.0004$ & $0.9640 \pm 0.0023$ & $0.9879 \pm 0.0289$ & $0.8536 \pm 0.0243$ & 65  & 87   & 7  \\
Times Square   & $0.0000 \pm 0.0000$ & $0.0000 \pm 0.0000$ & $0.0049 \pm 0.0003$ & $0.0000 \pm 0.0000$ & 242 & 421  & 16 \\
\bottomrule
\end{tabular}
\caption{Performance comparison of benchmark algorithms on real-world maps. The reported values represent the pursuer's worst-case utility at convergence, presented as mean $\pm$ standard deviation. The columns \textit{Nodes} and \textit{Edges} indicate the number of vertices and undirected connections in each graph, respectively. $T$ denotes the maximum time horizon for each game.}
\label{tab:real_world_benchmark}
\end{table*}

\subsection{Beyond Path-Based Evader Modeling}
\label{sec:beyond_path_based}

The scalability evaluation above shows that path enumeration is a major bottleneck for existing UNSG solvers. Meanwhile, GraphChase also supports step-wise policy execution, where agents sample actions directly from their policies at each decision point instead of committing to a complete path before the game starts. This allows agent strategies to be modeled as neural policies, such as PPO policies that make decisions step by step, or as rule-based heuristic policies. To validate this capability, we design two experiments on the Mumbai map, covering both unweighted and weighted settings.

We first consider the unweighted Mumbai map and train PPO policies for both the pursuer and the evader. Unlike the path-based evader protocol, the PPO evader selects actions step by step based on its observations during the game. After training, we evaluate the policies through cross-play with the pretrained PSRO agents. Table~\ref{tab:mumbai_ppo_crossplay} reports the defender capture rate over 1,000 simulations.

\begin{table}[t]
    \centering
    \caption{Cross-play evaluation on the unweighted Mumbai map. Each entry reports the defender capture rate over 1,000 simulations.}
    \label{tab:mumbai_ppo_crossplay}
    \begin{tabular}{lcc}
        \toprule
        \multirow{2}{*}{Attacker} & \multicolumn{2}{c}{Defender} \\
        \cmidrule(lr){2-3}
        & Pretrained PSRO & PPO \\
        \midrule
        Pretrained PSRO & 0.9790 & 0.5570 \\
        PPO             & 0.9400 & 0.7370 \\
        \bottomrule
    \end{tabular}
\end{table}

The results show that GraphChase can execute and evaluate step-wise PPO policies within the same environment. The pretrained PSRO defender achieves high capture rates against both attackers, while the PPO defender is less effective, especially against the pretrained PSRO attacker. This experiment confirms that the evader in GraphChase need not be represented only as a predetermined path, but can also be modeled as a step-wise reactive policy.

We further evaluate this policy-execution mode on a weighted Mumbai graph constructed from OpenStreetMap data. In this experiment, PPO policies are trained and evaluated directly on the weighted graph. We also include simple heuristic strategies: \textsc{Shortest-Path-to-Exit}, where the evader follows the exit with the minimum weighted shortest-path distance; \textsc{Risk-Aware Exit-Switching}, where the evader balances exit distance and pursuer proximity and may switch exits dynamically; and \textsc{Nearest-Threatened-Exit Guard}, where each pursuer moves toward the most threatened exit that it can reach fastest. Table~\ref{tab:mumbai_weighted_heuristic} reports the defender capture rate over 1,000 simulations.

\begin{table}[t]
    \centering
    \caption{Evaluation on the weighted Mumbai map. Each entry reports the defender capture rate over 1,000 simulations.}
    \label{tab:mumbai_weighted_heuristic}
    \begin{tabular}{lcc}
        \toprule
        Attacker / Defender & PPO & \shortstack{Nearest-threatened\\exit guard} \\
        \midrule
        PPO & 0.0200 & 0.0290 \\
        Shortest-path-to-exit & 0.0000 & 1.0000 \\
        Risk-aware exit-switching & 0.0010 & 1.0000 \\
        \bottomrule
    \end{tabular}
\end{table}

This experiment shows that GraphChase can train and evaluate policies directly on real weighted graphs, rather than only testing policies under randomly assigned edge costs. It also shows that heuristic strategies can be incorporated into the same execution and evaluation framework as neural policies. The heuristic defender captures the two heuristic attackers reliably, but performs poorly against the PPO attacker, suggesting that the learned attacker is less tied to simple exit-oriented behavior. Meanwhile, the PPO defender obtains low capture rates in the weighted setting, indicating that policy learning on real weighted urban graphs remains challenging.

\section{Related Works}
Game theory has emerged as a valuable tool in addressing   complex interactions and has been applied to security challenges \citep{Jain11,mccarthy2016preventing,sinha2018stackelberg}, including allocating limited resources to protect infrastructure \citep{Jain2013} or designing patrolling strategies in adversarial settings \citep{vorobeychik2014computing}. One important model is Stackelberg Security Games (SSGs), which is used to solve various security problems \citep{sinha2018stackelberg}. 
In SSGs, the defender moves first and then the attacker best responds to the defender's strategy.
UNSGs can be viewed as a networked variant of SSGs in zero-sum urban environments. Early work along this direction formulated urban security as game-theoretic resource allocation in networked domains \citep{tsai2010urban}, and later work improved UNSG scalability through cut-based graph contraction \citep{iwashita2016simplifying}.

Beyond this security-game perspective, UNSGs are also closely related to pursuit-evasion games \citep{parsons1976pursuit}, which involve strategic interactions between multiple pursuers and one or more evaders within a defined environment \citep{bilgin2015approach}. Pursuit-evasion games are classic research problems with applications in areas ranging from civilian safety \citep{oyler2016pursuit} to military operations \citep{vlahov2018developing}, and have been investigated in fields such as physics \citep{isaacs1965differential}, mathematics \citep{pachter1987simple, kopparty2005framework}, and engineering \citep{eklund2011switched}. Pursuit-evasion games were formulated as differential games, with foundational work by Rufus Isaacs in ``Differential Games'' \citep{isaacs1965differential}. Subsequent studies developed various algorithms, such as the linear–quadratic differential game (LQDG) approach by Ho et al. \citep{ho1965differential}, and the use of graphs to describe these problems by Parsons \citep{parsons1976pursuit}. Over time, the field expanded to include discrete settings, with work addressing discrete-time multiple-pursuer single-evader games \citep{bopardikar2008discrete}, and one-sided partially observable scenarios where the evader is fully informed, but the pursuers are not \citep{horak2016point, horak2017heuristic, horak2017dynamic}. Related domains such as patrolling security games (PSGs) have also been explored, where defenders guard against unseen intruders in stochastic, often infinite-horizon settings \citep{basilico2009leader, vorobeychik2014computing}. PSGs have been extended to accommodate uncertain signals and coordinated response \citep{basilico2017coordinating, basilico2017adversarial}. Our GraphChase can be extended to cover these settings. 

Existing multiplayer benchmarks for pursuit-evasion games, such as SIMPE~\citep{talebi2018simpe}, MARBLER~\citep{Jain11}, and Avalon~\citep{albrecht2022avalon}, have enriched the field with various testbeds. SIMPE explores diverse strategies with multiple pursuers and one evader, but operates in a continuous space and does not account for temporal constraints, which are essential in UNSGs. MARBLER links MARL with physical robots, while Avalon focuses on procedural worlds to evaluate generalization, primarily simulating biological survival rather than explicit pursuit-evasion. Similarly, environments like Google Research Football~\citep{kurach2020google} and Starcraft~\citep{samvelyan2019starcraft} provide MARL benchmarks in plane settings. However, these platforms emphasize algorithmic development and often overlook game-theoretic dynamics relevant to pursuit-evasion. Openspiel~\citep{lanctot2019openspiel} offers a wide range of games but does not include pursuit-evasion scenarios. As a result, a unified benchmark for modeling UNSGs remains lacking. Our GraphChase fills this gap by providing a flexible environment for UNSGs.

\begin{table}[tbp]
\centering
\small
\begin{tabular}{l|lcccc}
\toprule
 \multicolumn{1}{l}{} &  & Pretrained PSRO & Grasper & NSGZero & NSG-NFSP \\
\midrule
\multirow{2}{*}{ $7(0.5)$} & $T{=}4$ & $0.477$ & $0.481$ & $0.396$ & $0.466$ \\
                    & $T{=}8$ & $0.385$ & $0.236$ & $0.220$ & $0.107$ \\
\midrule
\multirow{2}{*}{ $7(1.0)$} & $T{=}4$ & $0.996$ & $0.989$ & $0.741$ & $0.982$ \\
                    & $T{=}8$ & $0.077$ & $0.195$ & $0.145$ & $0.267$ \\
\bottomrule
\end{tabular}
\captionof{table}{Robustness to horizon shift. The table shows the pursuer pseudo worst-case utility when the max time horizon is extended from $T=4$ to $T=8$.}
\label{tab:robustness}
\vspace{-8pt}
\end{table}

\section{Discussion:      Testbed for      Multiplayer Games   }

Computing an NE in multiplayer games is generally hard   \citep{chen20053,zhang2023computing}, and designing efficient algorithms for computing such an NE is still an open challenge. Our platform could be a testbed for algorithms for solving multiplayer games. 
The underlying graph-based pursuit-evasion and real-time interception structure is also relevant to other security domains, such as anti-poaching scenarios where UAVs can provide real-time location information of poachers \citep{bondi2018spot}. These domains share a common structure of graph-based pursuit, interception, and team coordination. In particular, GraphChase provides real-world scenarios for adversarial team games \citep{von1997team,basilico2017team,celli2018computational,farina2018ex,zhang2020computing,zhang2020converging,zhang2021computing,farina2021connecting,zhang2022correlation,zhang2022subgame,zhang2022optimal,zhang2022teamtree,carminati2022marriage,zhang2023team,mcaleer2023teampsro, Anagnostides2023Algorithms,li2023decision}, where a group of players competes against an adversary or another team. Various solution concepts apply depending on the situation. When team players compete independently against the adversary, the relevant solution concepts include 1) NE \citep{nash1951non,zhang2023computing}, where no player gains by deviating from this equilibrium, and 2) team-maxmin equilibrium (TME) \citep{von1997team,basilico2017team,celli2018computational,zhang2020computing,zhang2020converging,zhang2022correlation}, which is a type of NE that optimizes the team's utility across all NEs. Based on our platform, if we set that pursuers independently try to interdict the evader, we can also use our platform to compute an NE or TME in normal-form or extensive-form games. 
For normal-form games where team players can coordinate their strategies, the applicable solution concept is the correlated team-maxmin equilibrium (CTME) \citep{basilico2017team}. This is essentially equivalent to an NE in zero-sum two-player games, as the team with coordinated strategies and a unified payoff function behaves like a single player. 
In extensive-form games, the team with coordinated strategies has two solution concepts: 1) team-maxmin equilibrium with a communication device (TMECom) \citep{celli2018computational}, applicable when the team can continuously communicate and coordinate strategies, making the game akin to a two-player zero-sum game with perfect recall; and 2)   team-maxmin equilibrium with a coordination device (TMECor) \citep{celli2018computational,zhang2021computing,zhang2024dag}, used when the team can only coordinate strategies before gameplay, rendering the game similar to a two-player zero-sum game with imperfect recall. The algorithms in \citep{zhang2019optimal,li2021cfr,xue2021solving,xue2022nsgzero,li2023solving,li2024grasper} implemented on GraphChase compute a TMECom that is NE in team adversarial games. If we set that the team can only coordinate strategies before gameplay in extensive-form games, we can compute a TMECor on GraphChase. Future work may extend GraphChase from static edge weights to dynamic time-varying weights for modeling real-time urban traffic.

\textbf{Dual-use considerations.}
GraphChase is intended for defensive planning, benchmarking, and algorithmic evaluation for public-safety applications. However, it could in principle be misused to analyze escape strategies in insufficiently protected urban networks. We position GraphChase as a research and evaluation tool for defensive use, rather than an operational escape-planning system.

\begin{table}[tbp]
\small
\begin{tabular}{l|lccc}
\toprule
 \multicolumn{1}{l}{} &  & Pretrained PSRO & Grasper & NSG-NFSP \\
\midrule
\multirow{2}{*}{$7(0.5)$} & undirected graph & $0.483$ & $0.484$ & $0.435$ \\
                     & weighted graph   & $0.293$ & $0.343$ & $0.159$ \\
\midrule
\multirow{2}{*}{$7(1.0)$} & undirected graph & $0.983$ & $0.978$ & $0.679$ \\
                     & weighted graph   & $0.111$ & $0.259$ & $0.254$ \\
\bottomrule
\end{tabular}
\captionof{table}{Robustness to edge weight variations. Comparison of pursuer pseudo worst-case utility between unweighted undirected, and statically weighted version with the same connectivity. Edge weights are randomly sampled and fixed within each episode. NSGZero is excluded here as its online MCTS assumes unweighted rollouts and is not directly applicable to our weighted transfer setting.}
\label{tab:undirected_weighted}
\vspace{-18pt}
\end{table}

\section{Conclusion}

In this paper, we propose GraphChase, the first open-source platform dedicated to UNSGs, offering researchers a flexible multiplayer game environment and aiding in developing scalable algorithms. Specifically, we develop a unified and flexible platform to accommodate different UNSGs environments and implement several deep learning-based algorithms as benchmarks.
Extensive experiments on GraphChase provide insights into these algorithms.
We hope that GraphChase will facilitate the establishment of standardized evaluation criteria for UNSGs algorithms, thereby contributing to both theoretical advances and practical applications in solving general multi-agent games.


\textbf{Realistic Modeling:} Our platform stands out for its realistic modeling of complex, real-world environments, thanks to the advanced capabilities of GraphChase. By combining powerful graph-based representations with a hybrid decision-making mechanism, it effectively captures the dynamic and unpredictable nature of real scenarios — including fluctuating traffic patterns, spontaneous human behaviors, and other variable elements that traditional models often oversimplify. Furthermore, GraphChase provides native support for weighted graphs and user-updatable costs, enabling seamless integration of data-driven, time-dependent disruptions such as real-time traffic feeds, accident-induced road closures, or sudden congestion changes. This flexibility ensures that simulations remain highly faithful to actual conditions, delivering practical and reliable insights for real-world applications.

\begin{acks}
This research is supported by the InnoHK Funding and by the Ministry of Education, Singapore, under its Academic Research Fund Tier 1 (RG18/24).
\end{acks}

\bibliographystyle{ACM-Reference-Format}
\bibliography{reference}


\appendix

\section{Computational Scalability Analysis}
\label{app:scalability_analysis}

As discussed in the main text, current methods often rely on the exhaustive enumeration of the evader's action space. This requires pre-calculating all valid simple paths from the starting position to the exits within a finite time horizon $T$. While feasible for small maps, this approach faces severe scalability issues.

\begin{figure}[!htbp]
    \centering
    \includegraphics[width=0.85\columnwidth]{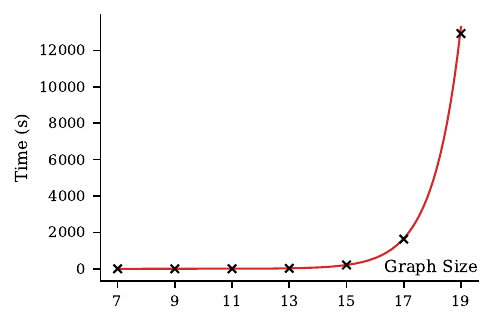}
    \caption{Time cost for enumerating evader simple paths from the grid center to four corners. The solid line represents the exponential trend, fitted via linear regression on the logarithmic scale. The rapid growth underscores the intractability of path-based methods on large-scale graphs.}
    \label{fig:attacker_time_consuming}
    \vspace{-10pt}
\end{figure}

To quantify this limitation, we measured the time cost required to enumerate simple paths from the center of a grid (N×N) to the corners, setting the time horizon $T=N$. Figure \ref{fig:attacker_time_consuming} illustrates the relationship between grid size and computation time. The results demonstrate an exponential explosion in computational overhead. Specifically, the runtime becomes prohibitively high once the grid size exceeds $20\times 20$. This trend confirms that reliance on global action space traversal constitutes a fundamental scalability barrier, necessitating the data-driven approach proposed in this work.

\section{User-controllable parameters}
\label{controllparams}



The user-controllable parameters are shown in Table \ref{controlableparams}. In our platform, users configure a graph-based pursuit-evasion game by providing a \textbf{base graph structure} and specifying game-specific settings through the \texttt{GameSettings} class. The base graph can be loaded from a NetworkX \texttt{gpickle} file, which defines the topology of the environment. We also provide a graph generation utility that allows users to create grid-based graphs through parameters such as \texttt{grid\_width} and \texttt{grid\_height}, which define the grid dimensions; \texttt{axis\_edge\_prob}, which controls the probability of retaining horizontal and vertical edges; and \texttt{diag\_edge\_prob}, which determines the probability of adding diagonal edges. These parameters enable users to generate diverse graph topologies ranging from sparse to fully-connected grid structures.

Users can further customize the graph structure through the \texttt{metadata} parameter. By specifying an \texttt{adjacency\_matrix} along with corresponding \texttt{node\_ids}, users can override the base graph topology with a completely custom structure. The adjacency matrix can be either \textbf{symmetric} (for undirected graphs) or \textbf{asymmetric} (for directed graphs), with each entry representing the edge weight between nodes. This flexible design enables users to model diverse scenarios, from simple grid-like environments to complex directed weighted networks. Additionally, the \texttt{use\_weighted\_graph} parameter controls whether edge weights are utilized in the game dynamics; when set to \texttt{False}, all edges default to unit weight.

For the \textbf{agent configuration and game parameters}, users control the number and initial positions of agents through \texttt{evader\_init} and \texttt{pursuer\_init}, which specify the starting node indices for each type of agent. The \texttt{exit\_nodes} parameter defines the set of goal locations that evaders aim to reach, while \texttt{time\_horizon} sets the maximum simulation duration. This parameterization allows users to construct a wide range of pursuit-evasion scenarios tailored to specific research questions or applications.
  
\begin{table}                                                                                                                       
\centering       
\caption{The parameters that users can control.}                                                                                    
\begin{tabular}{l|ll}                                                                                                               
\hline
&  \multicolumn{1}{c}{Graph Generation} & \multicolumn{1}{|c}{Graph Customization} \\
\hline
\multirow{4}{*}{%
  \begin{tabular}{@{}c@{}}
      underlying \\
      graph \\
      structure
  \end{tabular}
}
& \multicolumn{1}{c}{grid\_width} & \multicolumn{1}{|l}{adjacency\_matrix}\\
& \multicolumn{1}{c}{grid\_height} & \multicolumn{1}{|c}{node\_ids}\\
& \multicolumn{1}{c}{axis\_edge\_prob} & \multicolumn{1}{|c}{use\_weighted\_graph}\\
& \multicolumn{1}{c}{diag\_edge\_prob} & \multicolumn{1}{|l}{ }\\
\hline
\multirow{5}{*}{%
  \begin{tabular}{@{}c@{}}
      agent \\
      number \\
      and \\
      position
  \end{tabular}
}
& \multicolumn{2}{c}{time\_horizon} \\
& \multicolumn{2}{c}{evader\_init} \\
& \multicolumn{2}{c}{pursuer\_init} \\
& \multicolumn{2}{c}{exit\_nodes} \\
& \multicolumn{2}{c}{graph\_gpickle\_path} \\
\hline
\end{tabular}
\label{controlableparams}
\end{table}

\section{Faster wall-clock Convergence}
\label{convergefaster}

Our platform incorporates several architectural and implementation enhancements that contribute to faster wall-clock convergence times compared to the original implementations. These improvements span environment simulation, configuration management, data processing pipelines, and computational resource utilization.

\textbf{Standardized Environment Interface.} We have adopted the Gymnasium framework as the standard interface for all game environments, replacing the custom class-based implementations found in the original papers. The original code (e.g., \texttt{build\_game} in \texttt{game\_config.py}) constructs game instances through deeply nested function calls with numerous
embedded parameters, resulting in redundant data copying and inefficient state transitions. In contrast, our \texttt{UNSGEnv} class implements the standard Gymnasium API with optimized step and reset methods, eliminating unnecessary data transformations between the environment and agents. This standardization not only accelerates simulation processes but also ensures compatibility with modern reinforcement learning libraries.

\textbf{Efficient Graph Representation and Loading.} The original implementation rebuilds graph structures during environment initialization, often involving multiple graph transformations and node embedding computations for each experiment run. Our platform leverages NetworkX's \texttt{gpickle} serialization format to pre-compute and cache graph structures. Graphs are loaded once via \texttt{pickle.load()} and wrapped in a \texttt{GameSettings} object, which efficiently manages edge weights and neighbor mappings through pre-built dictionaries. This approach eliminates redundant graph construction overhead across training iterations.

\textbf{Streamlined Configuration Management.} We employ YAML-based configuration files combined with command-line overrides (via \texttt{load\_yaml\_config}), replacing the hardcoded parameter specifications scattered throughout the original codebase. The original implementation embeds configuration parameters directly in execution scripts, making parameter tuning and reproducibility challenging. Our configuration system centralizes all hyperparameters, enables programmatic parameter sweeps, and automatically saves experiment configurations via
\texttt{save\_experiment\_config}, facilitating efficient experimentation and result tracking.

\textbf{Optimized Data Processing Pipeline.} We have implemented various data processing optimizations throughout the training pipeline. The original code frequently converts between Python lists, numpy arrays, and PyTorch tensors, incurring significant overhead during high-frequency operations such as state observation extraction and action sampling. Our implementation maintains
consistent data representations using numpy arrays for observations (via \texttt{get\_current\_obs} returning structured numpy arrays) and PyTorch tensors for neural network computations, minimizing type conversion costs. Additionally, we have optimized the dimension computation functions (\texttt{compute\_input\_dim}, \texttt{compute\_action\_dim}) to perform calculations once during initialization rather than repeatedly during training.

\textbf{Modular Component Design.} Our platform separates concerns through well-defined abstractions: \texttt{env\_builder} for environment instantiation, \texttt{PPOAgent} for policy networks, \texttt{PPOAlgorithm} for learning updates, and specialized runners (\texttt{AttackerPathRunner}, \texttt{DefenderPretrainPsroRunner}) for training orchestration. This modularity contrasts with the monolithic \texttt{PSRO\_Oracle} class in the original implementation, which tightly couples environment interactions, policy updates, and meta-strategy computation. Our design reduces redundant computations by enabling component-level caching and reuse across PSRO iterations.

\textbf{Vectorized Environment Support.} Unlike the original sequential episode sampling, our platform supports vectorized environments through the \texttt{vec\_envs} parameter, enabling parallel data collection across multiple environment instances. This parallelization capability substantially reduces wall-clock training time, particularly during the pretrain phase where large numbers of episodes (\texttt{pretrain\_episodes\_per\_task}) are required.

It is important to emphasize that these enhancements primarily accelerate wall-clock convergence without altering the fundamental algorithmic behavior. The number of training iterations required for convergence remains approximately consistent with the original implementations. For instance, if the original PSRO algorithm requires $10^4$ episode samples to converge, our platform's reproduced version similarly requires a comparable number of training iterations. This consistency validates that our optimizations preserve algorithmic fidelity while delivering substantial wall-clock speedups through improved implementation efficiency.

\section{Description of Experimental Settings}
\label{appendix_games}
\begin{figure}[!htbp]
    \centering
    \begin{subfigure}[b]{0.45\textwidth}
        \centering
        \includegraphics[width=\textwidth]{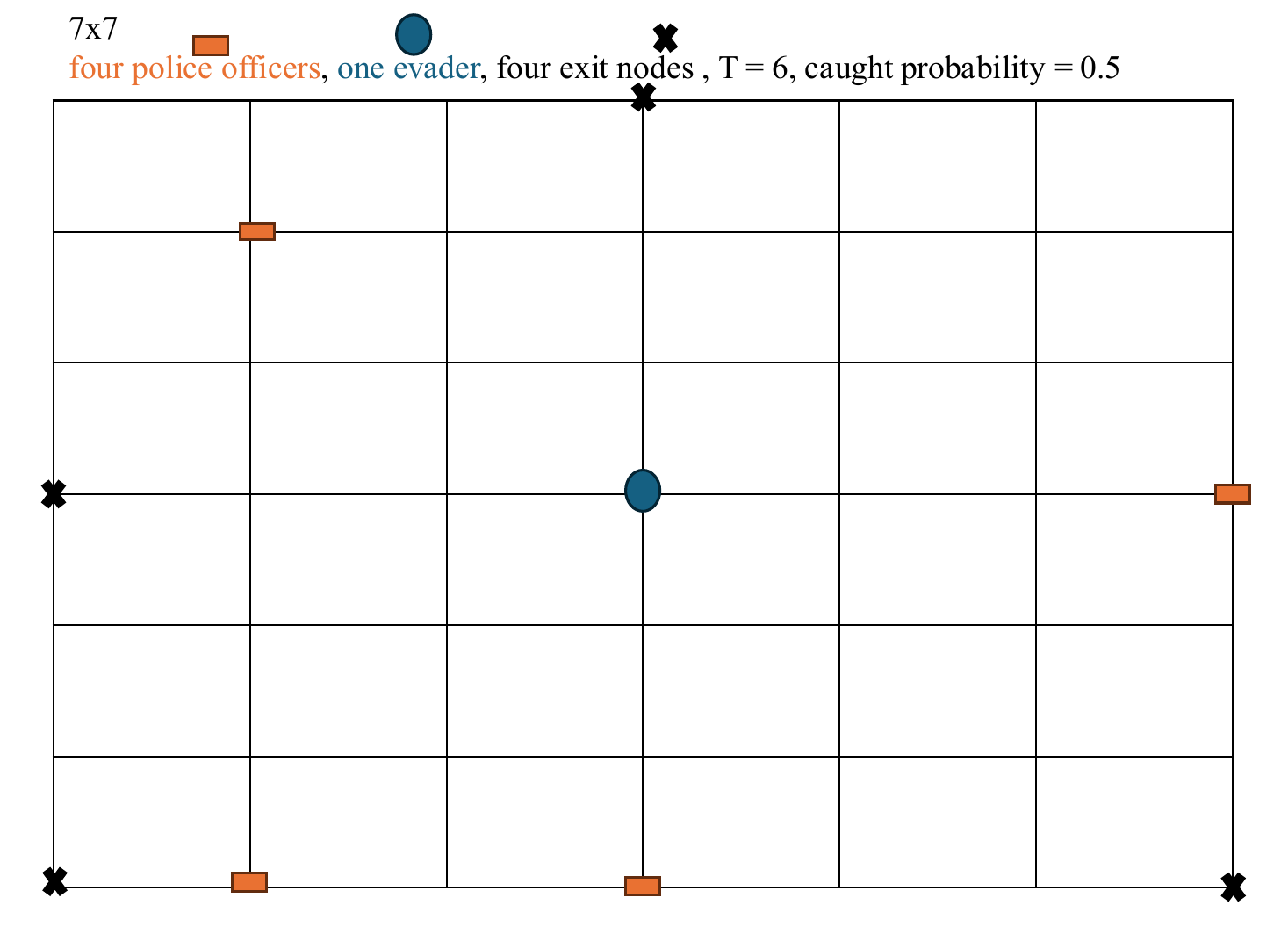}
    \end{subfigure}
    \begin{subfigure}[b]{0.45\textwidth}
        \centering
        \includegraphics[width=\textwidth]{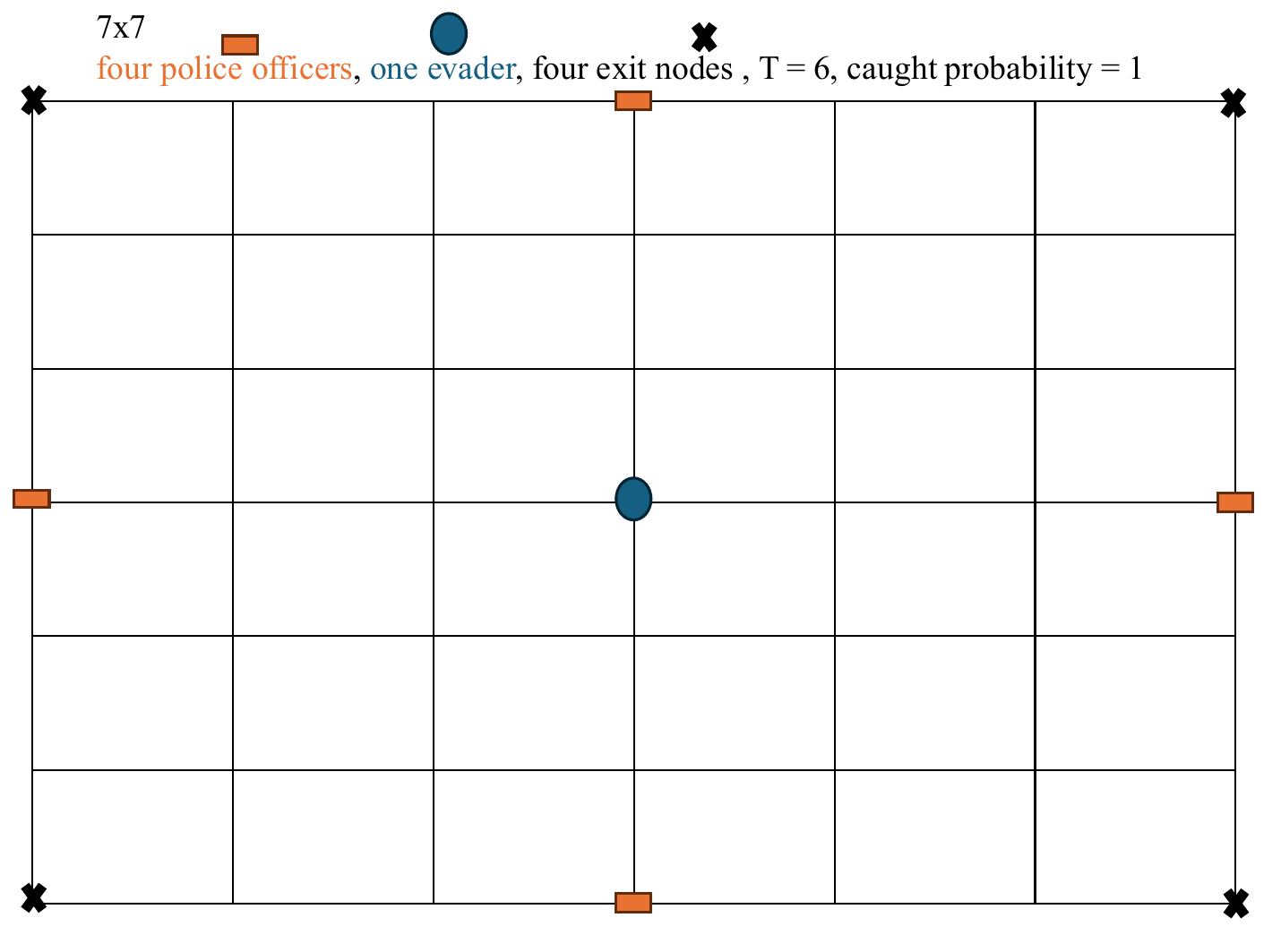}
    \end{subfigure}
    \caption{$7 \times 7$ grid graph settings with the evader caught probability of 0.5 (left) or 1 (right).}
    \label{game:7x7}
\end{figure}

\begin{figure}[!htbp]
    \centering
    \begin{subfigure}[b]{0.45\textwidth}
        \centering
        \includegraphics[width=\textwidth]{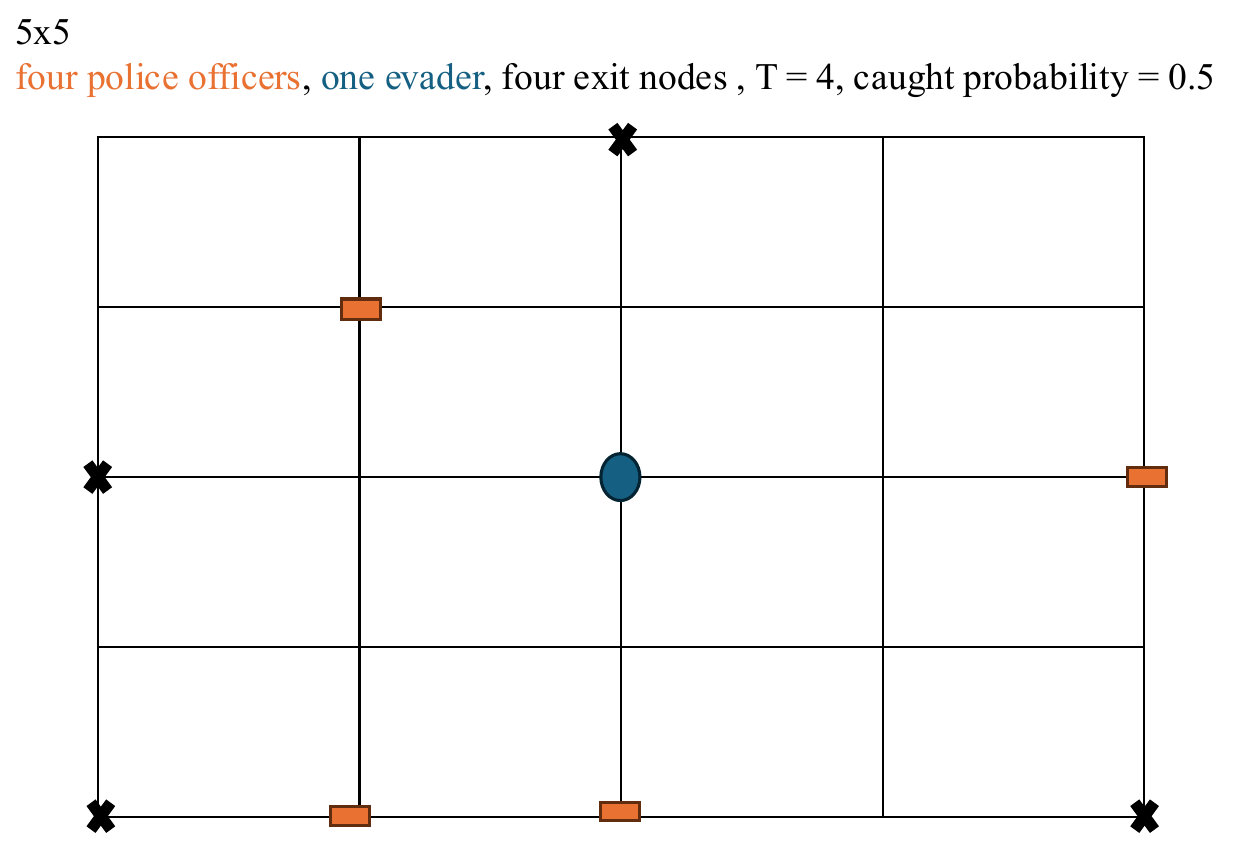}
    \end{subfigure}
    \begin{subfigure}[b]{0.45\textwidth}
        \centering
        \includegraphics[width=\textwidth]{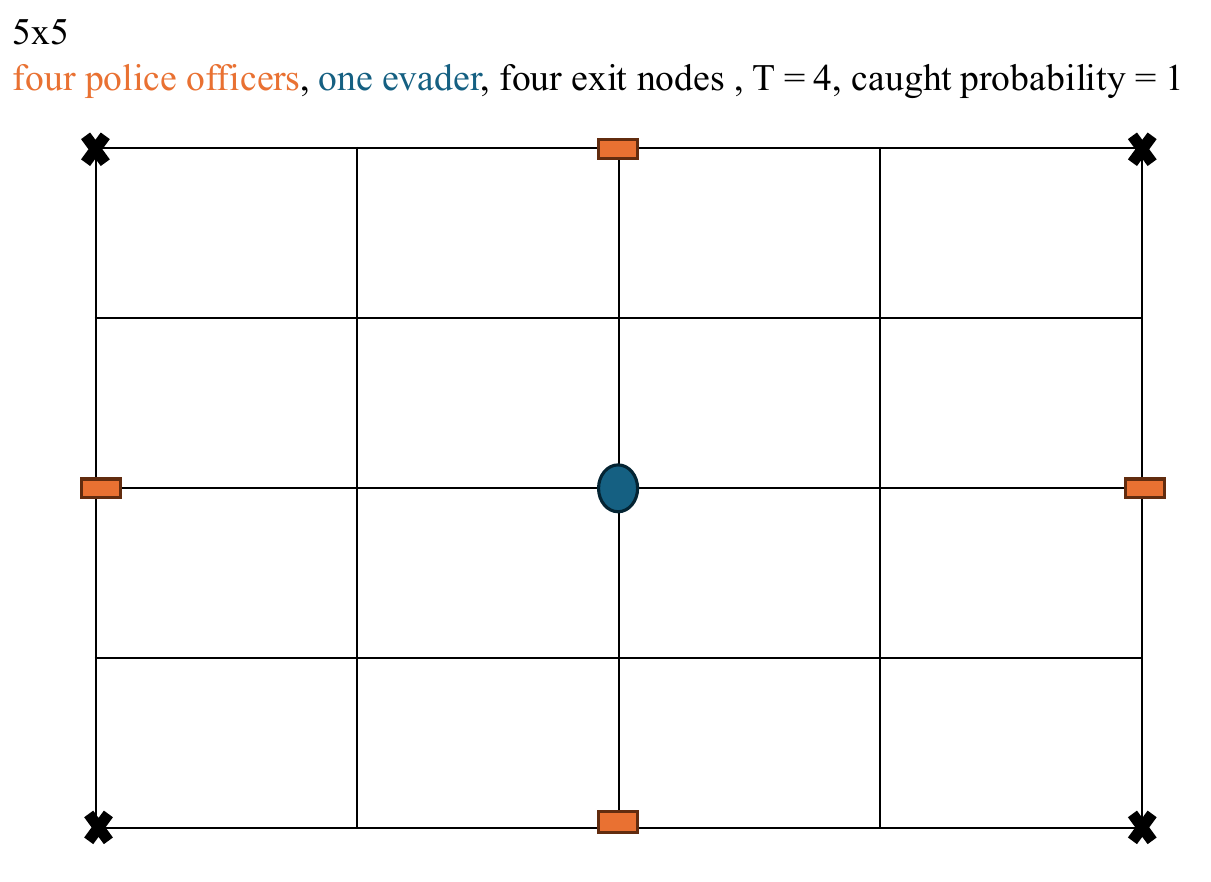}
    \end{subfigure}
    \caption{$5 \times 5$ grid graph settings with the evader caught probability of 0.5 (left) or 1 (right).}
     \label{game:5x5}
\end{figure}

The graph structures for the $5 \times 5$ and $7 \times 7$ grids are illustrated in Figure \ref{game:7x7} and \ref{game:5x5}. 

The reason why the left side of Figure \ref{game:5x5} reaches a probability of 0.5 under the Nash equilibrium is explained as follows:

\textbf{Exit Node Accessibility:}
Among the available exit nodes, two (located at the bottom-left and bottom-right) require more than three steps for the evader to reach. However, the pursuers can reach these exits in exactly three steps. As a result, a rational evader will avoid these exits, since choosing them would guarantee capture.

\textbf{Viable Exit Options:}
This constraint leaves only two viable exit nodes for the evader. Only the pursuer positioned at the top-left corner is able to reach either of these exits within three steps. Furthermore, both exits present equal strategic value for both the evader and the pursuer.

\textbf{Probability Calculation:}
Rational decision-making leads the evader to choose between these two viable exits with equal probability (i.e., $1/2$ for each). For each exit, the probability of being caught depends on whether the pursuer also chooses the same exit, which is again $1/2$. Thus, the total probability is determined as: $2$ exits $\times (1/2$ chance of the evader choosing each exit$) \times (1/2$ chance of the pursuer selecting the same exit$) = 0.5$.

A similar reasoning process applies to the scenario on the left side of Figure~\ref{game:7x7}. Under the Nash equilibrium, the probability that the evader is captured is also 0.5.

\section{Detailed Visualizations of Graph Structures}
\label{app:graph_structure_visualization}

\begin{figure*}[tbp]
    \centering
    \includegraphics[width=0.95\textwidth]{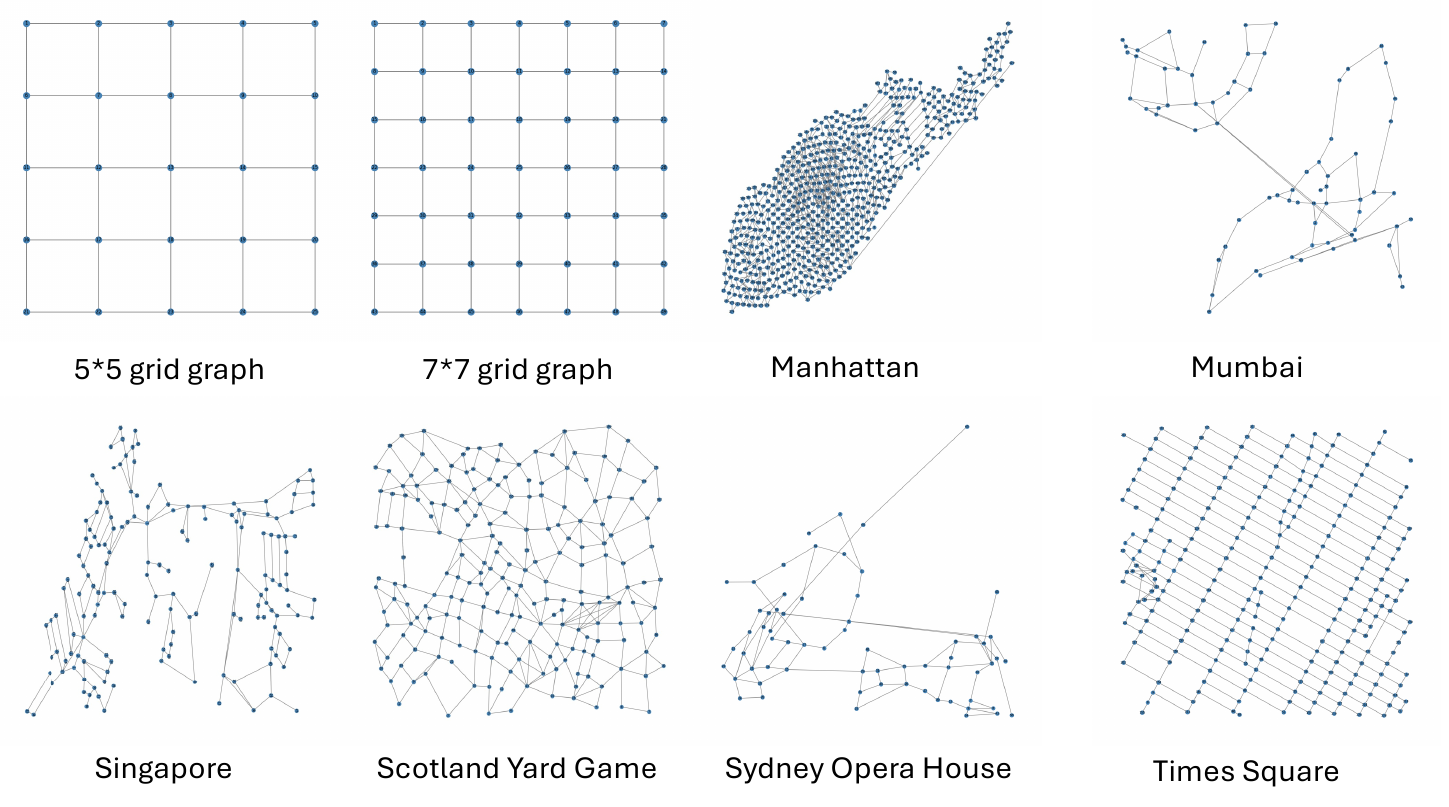}
    \caption{Visualization of the eight graph structures employed in the experiments. The dataset comprises synthetic grid graphs ($5\times 5, 7\times 7$) and real-world topologies.
    }
    \label{fig:exp_graph_structures}
\end{figure*}

To provide further details on the experimental setup, we present the specific network topologies used in our evaluation. Figure \ref{fig:exp_graph_structures} illustrates the eight distinct graph structures employed. These include both structured synthetic grids and irregular real-world road networks.

\section{Visualization Evaluation}
\label{sec:visualization}

\begin{figure*}[tbp]
    \centering
    \includegraphics[width=0.9\textwidth]{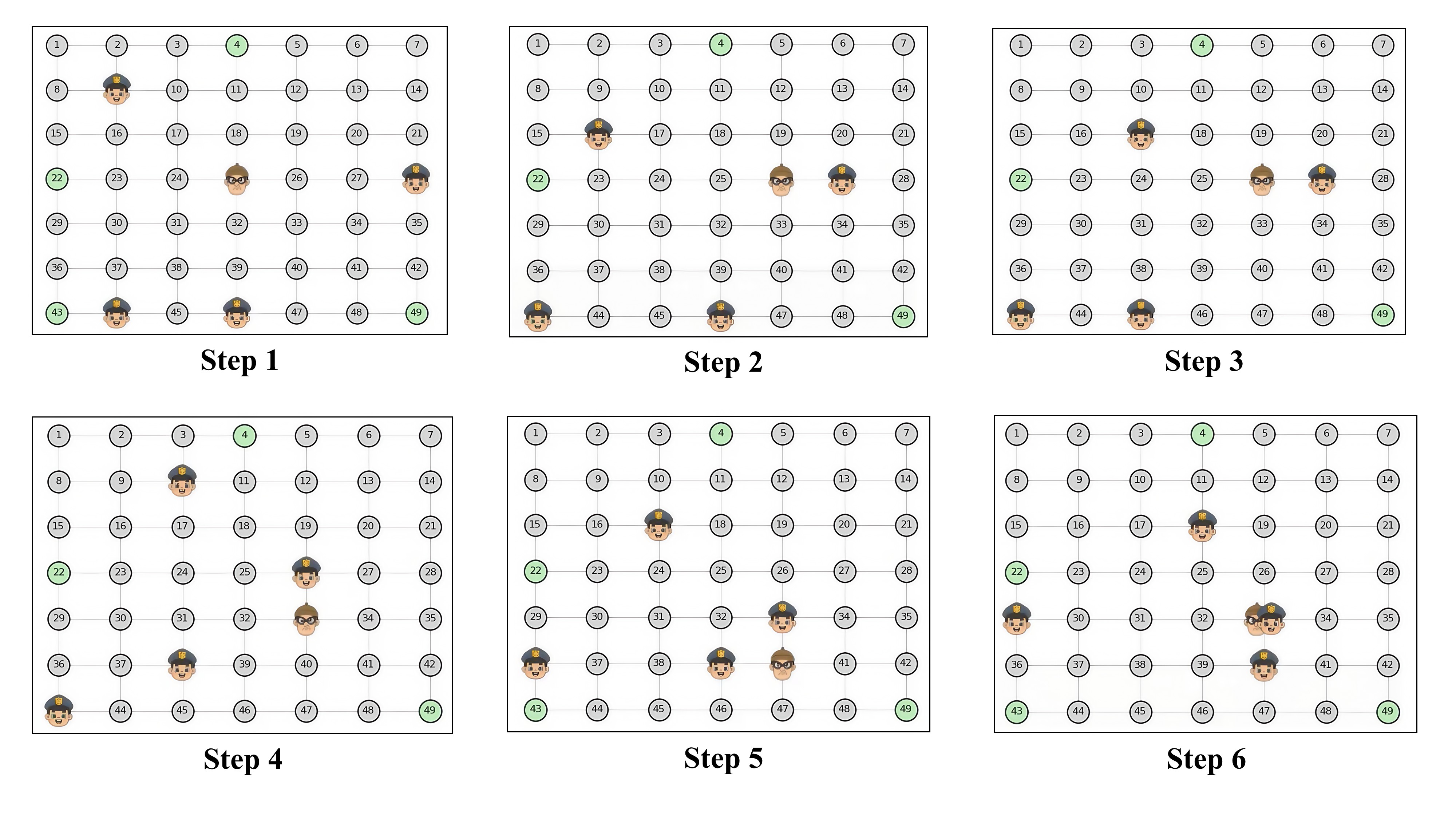}
    \caption{Snapshots from a recorded pursuit-evasion episode.}
    \label{fig:visualization}
\end{figure*}

This section presents the Visualization evaluation method provided by GraphChase. Figure \ref{fig:visualization} captures snapshots from a recorded pursuit-evasion episode, illustrating the position changes of the pursuers and the evader at each time step. Step 1 shows the initial positions of the evader and the pursuers. Each subsequent step reflects one move by each side. In Step 6, the evader and pursuer overlap, indicating that the evader has been captured by a pursuer.

\section{Usage Instructions for GraphChase}
\label{usage}

The following steps outline the process for setting up and utilizing the GraphChase platform:                                       

\subsection{Cloning the Repository}
To begin, clone the GraphChase repository from GitHub and navigate to the project directory:
\begin{verbatim}
git clone \
https://github.com/GraphChase/GraphChasePlatform.git
cd GraphChasePlatform
\end{verbatim}

\subsection{Installing Dependencies}
Install the necessary dependencies, including PyTorch, NetworkX, Gymnasium, and other required packages. Refer to the
\nolinkurl{requirements.txt} file for the complete list of dependencies.

\subsection{Preparing Graph Structures}
Our platform uses NetworkX \texttt{gpickle} format for graph storage. Pre-generated graph files are available in the
\nolinkurl{new\_version/graph/custom\_graph/} directory. To create custom graphs, use the graph generation utility:
\begin{verbatim}
python -m new_version.graph.generate_custom_graph \
  --grid_width 7 --grid_height 7 \
  --axis_edge_prob 1.0 --diag_edge_prob 0.0 \
  --output_path custom_graph.gpickle
\end{verbatim}

\subsection{Running an Algorithm}
To run a specific algorithm, such as Pretrain-PSRO, perform the following steps:

\begin{enumerate}[left=0pt,labelsep=5pt]
  \item \textbf{Select or Create Configuration File:} Choose an appropriate YAML configuration file from the
\texttt{new\_version/solver\_cfgs/} directory (e.g., \texttt{pretrain\_psro\_cfgs\_7\_7\_5.yaml}), or create a new one based on
existing templates. The configuration file specifies:
  \begin{itemize}
      \item \textbf{Graph arguments:} Path to the gpickle file (\nolinkurl{graph\_gpickle\_path}), agent initial positions
(\nolinkurl{attacker\_init}, \nolinkurl{defender\_init}), exit nodes (\texttt{exit\_nodes}), and time horizon
      \item \textbf{Graph embedding arguments:} Whether to use graph embeddings, embedding parameters, and paths to pre-computed
embeddings
      \item \textbf{Algorithm hyperparameters:} PSRO iterations, training batch sizes, PPO learning rates, and other
algorithm-specific parameters
      \item \textbf{Experiment settings:} Random seed, GPU configuration, output paths, and logging preferences
  \end{itemize}

  \item \textbf{Customize Parameters (Optional):} To override specific configuration values without modifying the YAML file, use
command-line arguments with the \texttt{--set} flag. For example, to change the random seed:
  \begin{verbatim}
  --set seed=3407
  \end{verbatim}

  \item \textbf{Run the Algorithm:} Execute the algorithm using one of the following methods:

  \textbf{Method 1 - Direct Python Execution:}
  \begin{verbatim}
  python -m new_version.scripts.run_pretrain_psro \
      --config new_version/solver_cfgs/\
pretrain_psro_cfgs_7_7_5.yaml \
      --set seed=3407
  \end{verbatim}

  \textbf{Method 2 - Shell Script for Multi-seed Experiments:}
  \begin{verbatim}
  bash run_pretrainpsro_multiseed.sh
  \end{verbatim}
  This script automates the process of running experiments across multiple random seeds, particularly useful for statistical validation and cluster computing environments.
\end{enumerate}






All algorithms utilize the same YAML-based configuration system, enabling consistent parameterization and reproducible
experimentation across different solution methods.

\section{Hyperparameter Configuration}                                               \label{hyperparams}

The hyperparameter configurations for all implemented algorithms are detailed in Tables~\ref{tab:pretrain_psro_hyperparams},
\ref{tab:cfrmix_hyperparams}, \ref{tab:nsgzero_hyperparams}, \ref{tab:nsgnfsp_hyperparams}, and \ref{tab:grasper_hyperparams}. These
configurations are based on the experimental setting with a $7 \times 7$ grid graph. For additional experimental configurations, including different graph sizes, agent numbers, and real-world map scenarios, pre-configured YAML files are
available in the \texttt{new\_version/solver\_cfgs/} directory of our GitHub repository.

\begin{table*}[t]
\centering
\caption{Hyperparameters for Pretrain-PSRO algorithm (7×7 grid, captured $prob=0.5$).}
\label{tab:pretrain_psro_hyperparams}
\small
\begin{tabular}{ll|ll}
\hline
\multicolumn{4}{c}{\textbf{Game Configuration}} \\
\hline
time\_horizon & 7 & seed & 77 \\
use\_weighted\_graph & false & device\_id & 0 \\
\hline
\multicolumn{4}{c}{\textbf{Graph Embedding Parameters}} \\
\hline
graph\_embeddings & true & emb\_size & 16 \\
node\_information\_type & all & normalize\_info & true \\
similarity & cosine & line\_order & all \\
epochs & 200 & batch\_size & 32 \\
neg\_samples & 5 & lr & 0.025 \\
\hline
\multicolumn{4}{c}{\textbf{PSRO Parameters}} \\
\hline
num\_psro\_iteration & 20 & eval\_episodes & 1000 \\
rollouts\_per\_attacker\_action & 10000 & vec\_envs & 16 \\
train\_defender\_batches & 630 & episodes\_per\_batch & 8 \\
\hline
\multicolumn{4}{c}{\textbf{Pretraining Parameters}} \\
\hline
pretrain\_iterations & 40 & pretrain\_tasks & 30 \\
pretrain\_episodes\_per\_task & 20 & load\_pretrained\_model & false \\
\hline
\multicolumn{4}{c}{\textbf{PPO Hyperparameters}} \\
\hline
ppo\_actor\_lr & 0.001 & ppo\_critic\_lr & 0.003 \\
ppo\_gamma & 0.99 & ppo\_lambda & 0.95 \\
ppo\_clip & 0.2 & ppo\_epochs & 10 \\
ppo\_batch\_size & 32 & ppo\_hidden\_dim & 128 \\
entropy\_coef & 0.01 & & \\
\hline
\multicolumn{4}{c}{\textbf{Attacker Settings}} \\
\hline
action\_type & exit\_node & strategy\_type & mix \\
attacker\_path\_type & shortest & reward\_mode & win\_rate \\
\hline
\end{tabular}
\end{table*}

\begin{table*}[t]
\centering
\caption{Hyperparameters for CFR-Mix algorithm (7×7 grid, captured $prob=0.5$).}
\label{tab:cfrmix_hyperparams}
\small
\begin{tabular}{ll|ll}
\hline
\multicolumn{4}{c}{\textbf{Game Configuration}} \\
\hline
time\_horizon & 7 & seed & 77 \\
use\_weighted\_graph & false & device\_id & 0 \\
\hline
\multicolumn{4}{c}{\textbf{Network Architecture}} \\
\hline
network\_dim & 32 & & \\
\hline
\multicolumn{4}{c}{\textbf{Training Parameters}} \\
\hline
iteration & 10000 & train\_epoch & 1000 \\
sample\_number & 20 & action\_number & 1000 \\
\hline
\multicolumn{4}{c}{\textbf{Attacker Regret Network}} \\
\hline
attacker\_regret\_batch\_size & 32 & attacker\_regret\_lr & 0.0015 \\
\hline
\multicolumn{4}{c}{\textbf{Defender Regret Network}} \\
\hline
defender\_regret\_batch\_size & 512 & defender\_regret\_lr & 0.0015 \\
\hline
\multicolumn{4}{c}{\textbf{Defender Strategy Network}} \\
\hline
defender\_strategy\_batch\_size & 32 & defender\_strategy\_lr & 0.0015 \\
\hline
\end{tabular}
\end{table*}

\begin{table*}[t]
\centering
\caption{Hyperparameters for NSGZero algorithm (7×7 grid, captured $prob=0.5$).}
\label{tab:nsgzero_hyperparams}
\small
\begin{tabular}{ll|ll}
\hline
\multicolumn{4}{c}{\textbf{Game Configuration}} \\
\hline
time\_horizon & 7 & seed & 77 \\
use\_weighted\_graph & false & device\_id & 0 \\
save\_model & true & use\_tensorboard & false \\
\hline
\multicolumn{4}{c}{\textbf{Network Architecture}} \\
\hline
embedding\_dim & 16 & hidden\_dim & 256 \\
\hline
\multicolumn{4}{c}{\textbf{Training Parameters}} \\
\hline
max\_episodes & 6000 & batch\_size & 128 \\
buffer\_size & 50000 & lr & 0.0005 \\
num\_workers & 4 & debug\_single\_process & false \\
\hline
\multicolumn{4}{c}{\textbf{Training Schedule}} \\
\hline
train\_every & 16 & train\_from & 128 \\
test\_every & 250 & test\_nepisodes & 50 \\
save\_every & 500 & log\_every & 500 \\
\hline
\multicolumn{4}{c}{\textbf{MCTS Parameters}} \\
\hline
num\_sims & 150 & bias & 0.5 \\
cpuct & 0.3 & temp & 0.5 \\
gamma & 1.0 & reward\_mode & win\_rate \\
\hline
\multicolumn{4}{c}{\textbf{Attacker Configuration}} \\
\hline
att\_type & nfsp & ban\_capacity & 500 \\
cache\_capacity & 20 & br\_rate & 0.2 \\
\hline
\end{tabular}
\end{table*}

\begin{table*}[t]
\centering
\caption{Hyperparameters for NSG-NFSP algorithm (7×7 grid, captured $prob=0.5$).}
\label{tab:nsgnfsp_hyperparams}
\small
\begin{tabular}{ll|ll}
\hline
\multicolumn{4}{c}{\textbf{Game Configuration}} \\
\hline
time\_horizon & 7 & seed & 0 \\
use\_weighted\_graph & false & device\_id & 0 \\
\hline
\multicolumn{4}{c}{\textbf{Network Architecture}} \\
\hline
embedding\_size & 32 & hidden\_size & 64 \\
relevant\_v\_size & 64 & if\_naivedrrn & false \\
seq\_mode & cnn & & \\
\hline
\multicolumn{4}{c}{\textbf{Replay Buffer Configuration}} \\
\hline
br\_buffer\_capacity & 500000 & avg\_buffer\_capacity & 10000000 \\
\hline
\multicolumn{4}{c}{\textbf{Learning Parameters}} \\
\hline
br\_lr & 0.0001 & avg\_lr & 0.0001 \\
br\_batch\_size & 128 & avg\_batch\_size & 256 \\
\hline
\multicolumn{4}{c}{\textbf{Exploration Parameters}} \\
\hline
d\_expl & 0.0 & a\_expl & 0.1 \\
br\_prob & 0.1 & & \\
\hline
\multicolumn{4}{c}{\textbf{Training Schedule}} \\
\hline
max\_episodes & 1200000 & train\_br\_freq & 4 \\
train\_avg\_freq & 32 & check\_freq & 100000 \\
check\_from & 100000 & display\_freq & 10000 \\
min\_to\_train & 1000 & exact\_br & false \\
\hline
\multicolumn{4}{c}{\textbf{Agent Configuration}} \\
\hline
defender\_rl\_mode & drrn & defender\_sl\_mode & drrn \\
attacker\_mode & bandit & reward\_mode & win\_rate \\
\hline
\end{tabular}
\end{table*}

\begin{table*}[t]
\centering
\caption{Hyperparameters for GRASPER-MAPPO algorithm (7×7 grid, captured $prob=0.5$).}
\label{tab:grasper_hyperparams}
\small
\begin{tabular}{ll|ll}
\hline
\multicolumn{4}{c}{\textbf{Game Configuration}} \\
\hline
time\_horizon & 7 & seed & 77 \\
use\_weighted\_graph & false & device\_id & 0 \\
graph\_type & Grid\_Graph & edge\_probability & 0.8 \\
row & 7 & column & 7 \\
\hline
\multicolumn{4}{c}{\textbf{Graph Embedding Configuration}} \\
\hline
use\_node\_emb & true & load\_graph\_emb\_model & true \\
max\_epoch & 200 & & \\
\hline
\multicolumn{4}{c}{\textbf{Game Pool Configuration}} \\
\hline
load\_game\_pool\_file & true & pool\_size & 1000 \\
min\_num\_defender & 4 & max\_num\_defender & 4 \\
min\_num\_exit & 4 & max\_num\_exit & 5 \\
min\_attacker\_pth\_len & 5 & & \\
\hline
\multicolumn{4}{c}{\textbf{PSRO Parameters}} \\
\hline
num\_psro\_iteration & 20 & train\_defender\_batches & 1200 \\
episodes\_per\_batch & 32 & reward\_mode & win\_rate \\
\hline
\multicolumn{4}{c}{\textbf{MAPPO Hyperparameters}} \\
\hline
ppo\_epoch & 8 & minibatch\_size & 64 \\
use\_gae & false & use\_act\_supervisor & true \\
\hline
\multicolumn{4}{c}{\textbf{Training Schedule}} \\
\hline
num\_iterations & 20000 & save\_every & 1000 \\
\hline
\end{tabular}
\end{table*}

\end{document}